\documentclass[%
    10pt,
 superscriptaddress,
 nofootinbib,
 amsmath,amssymb,
 aps,
 pre
]{revtex4-2}

\usepackage{graphicx}
\usepackage{dcolumn}
\usepackage{bm}
\usepackage{xcolor}
\usepackage[ruled,lined]{algorithm2e}
\usepackage{physics} 
\usepackage{dsfont}
\usepackage{comment}

\usepackage[
margin=1.2in,
]{geometry}

\newcommand{\vect}[1]{\boldsymbol{\mathbf{#1}}}
\renewcommand{\eqref}[1]{\textup{{\normalfont Eq.~(\ref{#1}}\normalfont)}}
\newcommand{\giventhat}[2]{(\,#1\mid#2\,)}

\DeclareMathOperator{\argmax}{arg\,max}

\begin{document}


\title{Inference in Spreading Processes with Neural-Network Priors
}
\author{Davide Ghio}%
\affiliation{%
 Information, Learning and Physics Laboratory. Ecole Polytechnique Fédérale de Lausanne (EPFL). 
}%
\author{Fabrizio Boncoraglio}
\affiliation{%
 Statistical Physics of Computation Laboratory. Ecole Polytechnique Fédérale de Lausanne (EPFL). 
}%
\author{Lenka Zdeborová}
\affiliation{%
 Statistical Physics of Computation Laboratory. Ecole Polytechnique Fédérale de Lausanne (EPFL).
}%


\begin{abstract}
 
Stochastic processes on graphs are a powerful tool for modelling complex dynamical systems such as epidemics. A recent line of work focused on the inference problem where one aims to estimate the state of every node at every time, starting from partial observation of a subset of nodes at a subset of times. In these works, the initial state of the process was assumed to be random i.i.d. over nodes. Such an assumption may not be realistic in practice, where one may have access to a set of covariate variables for every node that influence the initial state of the system. In this work, we will assume that the initial state of a node is an unknown function of such covariate variables. Given that functions can be represented by neural networks, we will study a model where the initial state is given by a simple neural network -- notably the single-layer perceptron acting on the known node-wise covariate variables.

Within a Bayesian framework, we study how such neural-network prior information enhances the recovery of initial states and spreading trajectories. We derive a hybrid belief propagation and approximate message passing (BP-AMP) algorithm that handles both the spreading dynamics and the information included in the node covariates, and we assess its performance against the estimators that either use only the spreading information or use only the information from the covariate variables.
 
We show that in some regimes, the model can exhibit first-order phase transitions when using a Rademacher distribution for the neural-network weights. These transitions create a statistical-to-computational gap where even the BP-AMP algorithm, despite the theoretical possibility of perfect recovery, fails to achieve it. 
\end{abstract}

\maketitle


\section{Introduction}
Spreading  dynamics on networks model many important systems, such as information diffusion in social networks~\cite{vespignani2012modelling}, epidemic processes~\cite{pastor2015epidemic,keeling2005networks}, and gene regulatory networks~\cite{karlebach2008modelling}. There are many studies in the literature that tackle the interplay between the network topology and the dynamical process which takes place on top of it~\cite{boccaletti2006complex}.

In this work we study Bayesian inference for this class of models, aiming to recover some information about a realisation of the dynamics from partial observations. In such Bayesian setting, this problem reduces to computing posterior expectations, where the spreading process itself defines the prior and observations form the likelihood.
Statistical physics has yielded several inference algorithms for spreading models on random sparse graphs, including dynamic message passing (DMP)~\cite{lokhov2014inferring} and belief propagation (BP)~\cite{altarelli2014bayesian}. Our work builds on the BP formulation to derive a new algorithm.

Spreading models often assume a separable prior over initial sources, ignoring node-specific information. However, node features are often available in practice and can enhance inference. To leverage such data, recent work in signal processing uses generative priors, often based on neural networks that map features to problem variables~\cite{aubin2019spiked,aubin2020exact}.

We introduce the Neural Sources Spreading (NSS) model, where the unknown initial state of the spreading process is given by the output of a neural network with unknown weights acting on known node-wise variables. The NSS model's posterior corresponds to a hybrid factor graph. It combines a sparse graph on which the spreading dynamics takes place - suited for BP on locally tree-like structures~\cite{altarelli2014bayesian,mezard2009information} - with a dense graph for the neural-network prior, which is effectively handled by approximate message passing (AMP) in high dimensions~\cite{zdeborova2016statistical,barbier2019optimal}. Using the cavity method, we merge these components to derive a hybrid BP-AMP algorithm that handles both the sparse spreading interactions and the dense Neural-Network prior~\cite{manoel2017multi,duranthon2023neural}.

This paper is structured as follows. Section~\ref{sec:Model} defines the NSS model. Section~\ref{sec:BI} details the Bayesian inference framework and performance metrics. Section~\ref{sec:BPAMP} presents the cavity equations and our BP-AMP algorithm. We then analyse its performance in a setting showing evidence of Bayes-optimality (Sec.~\ref{sec:Gauss}) and another exhibiting a statistical-to-computational gap (Sec.~\ref{sec:Rade}). 
The code to reproduce all results is available in a dedicated GitHub repository\footnote{\url{https://github.com/IdePHICS/NeuralSpreadingInference}}.

\subsection{Related works}
This study is situated within the field of spreading processes on graphs, which examines how these dynamics relate to the underlying graph structure, particularly in domains such as epidemiology~\cite{newman2003structure,kiss2017mathematics}. Building on established compartmental models~\cite{Kermack1932,Pare2020}, we focus on the algorithmic challenges of inference, such as identifying the initial sources of an epidemic or recovering entire infection trajectories.

A large body of work in different scientific domains have studied the problem of inference of spreading processes on graphs. Probably the most important example is the one of source detection, or equivalently of finding the patient-zero of the epidemic, which has attracted many researchers after the seminal papers~\cite{Shah2010,Shah2011}, where the authors, restricting to the SI model, showed rigorous results for regular trees using the concept of rumour centrality as maximum likelihood estimator. Later, in~\cite{AntulovFantulin2014} the authors extended the study to general trees, and in~\cite{dong2013rooting} the SIR model was considered, in both cases choosing as partial information a snapshot of the epidemic, i.e.\ the state of all the nodes at a particular time. At the same time in~\cite{Pinto2012} the authors moved to consider localised observations, or sensors, as source of information and described the optimal maximum likelihood estimator for trees.

In~\cite{Karrer2010} for the first time a message-passing approach was introduced to analyse epidemic models, enabling the computation of average properties on arbitrary random graphs ensembles. Later, two different works were able to design a message-passing algorithm operating on single graphs: in~\cite{lokhov2014inferring} the authors introduced a ``Dynamical Message Passing'' (DMP) algorithm, which is exact on trees but gives results for a fixed initial condition, while in~\cite{altarelli2014patient} the ``Belief Propagation'' (BP) equations for this problem were derived for the first time, and were used to infer the origin of epidemics with multiple patients zero and various types of partial information. These techniques were later used for other related problems, as the reconstruction of the model's parameters~\cite{lokhov2016reconstructing} or epidemic mitigation studies~\cite{baker2021epidemic}.

The methodology we employ connects to the idea of merging graphical models, an approach that was formulated in~\cite{manoel2017multi,gabrie2018entropy} to extend single-layer AMP to multiple layers. Closer to the present setting,~\cite{duranthon2023neural} derived a BP-AMP algorithm for community detection with a neural network prior on node attributes.

Finally, we note a recent line of work conjecturing the existence of replica symmetry breaking in spreading processes at very low source densities~\cite{braunstein2023statistical,braunstein2025evidence}. This paper, in contrast, focuses on scenarios with higher source densities where this phenomenology is not observed, leaving the low-density regime with a neural-network prior for future study.

\section{\label{sec:Model}The Neural Sources Spreading Model}
\subsection{Spreading models on graphs}
We consider processes on a graph $G(V,E)$ with $N$ nodes.
Each node of the graph is assigned a variable $x_i^t$ at discrete times $t\in\{0,1,\dots,T\}$, taking values among a finite number $p$ of possible states. 
We focus on unidirectional dynamics, where a node's transition times between the possible states of the nodes  $\{\vect{t_i}\}_{i=1}^N = \{t^1_i,\dots,t_i^{p-1}\}_{i=1}^N$, which we define as the last time in which a node is in a state before it transitions, fully specify its evolution. This static representation reduces the system's variables from $O(N \times T)$ to $O(N)$. This restriction can be relaxed to models where nodes revisit states (e.g., SIS) by treating backward steps as new state transitions~\cite{ghio2023bayes,pastor2015epidemic}.

Assuming local spreading dynamics, where transitions depend only on nearest-neighbour states, the transition time probability factorizes as:
\begin{equation}\label{eq:Spread1}
    P\giventhat{\{\vect{t_i}\}_{i=1}^N}{ \vect{\Theta}} = \frac{1}{Z_{\rm spread}} \prod_{i=1}^N \Psi_i(\vect{t_i},\{\vect{t_j}\}_{j\in\partial_i}, \vect{\Theta})\,,
\end{equation}
where $Z_{\rm spread}$ is the normalization, $\vect{\Theta}$ are model parameters, and $\partial_i$ is the set of neighbours of node $i$. Importantly, the factor $\Psi_i (\vect{t_i},\{\vect{t_j}\}_{j\in\partial_i}, \vect{\Theta} )$ is a node factor which depends on its transition time and the ones of its nearest neighbours. 

We will illustrate the method on two examples of the dynamical model: the Susceptible-Infected (SI) model and the deterministic Susceptible-Infected-Recovered (dSIR) model~\cite{ghio2023bayes}. 
\begin{itemize}
    \item In the SI model, each node is either susceptible (S) or infectious (I), so the dynamics is described by a single transition time $t_i$ for each node $i$. We say that node $i$ is a \textit{source} of the spreading if it is infectious at time $t=0$, and in this case we have $t_i = -1$.  At time $t$, a susceptible node $i$ has a probability $\lambda_{ji}^t\in [0,1]$ to get infected from an infectious node $j\in\partial_i$. 
    In mathematical terms, the probability for node $i$ of being infected at time $s$ is given by
    \begin{equation}
        {p}^{\rm inf}_i(s,  \{t_k\}_{k\in\partial_i}) = 1 -\prod_{k\in \partial_i} (1-\lambda_{ki}\mathds{I}[s>t_k])\,.
    \end{equation}
    The total factor for this node is then
    \begin{equation}\label{eq:FactSI}
        \Psi_i\giventhat{t_i, \{t_k\}_{k\in\partial_i}}{x_i^0} = 
        \begin{cases}
        {p}^{\rm inf}_i(t_i,  \{t_k\}_{k\in\partial_i}) \prod_{s=0}^{t_i-1}  \left( 1 - {p}^{\rm inf}_i(s,  \{t_k\}_{k\in\partial_i}) \right) & \text{if }x_i^0 = S,\ t_i<T \\ 
        \prod_{s=0}^{T-1}  \left( 1 - {p}^{\rm inf}_i(s,  \{t_k\}_{k\in\partial_i}) \right) & \text{if }x_i^0 = S,\ t_i=T \\ 
        \delta_{t_i,-1} & \text{if }x_i^0 = I\,,
        \end{cases}
    \end{equation}
    where we assign $t_i = T$ to nodes which are still susceptible at $t=T-1$, the last simulation time.
    \item In the dSIR model, S-to-I transitions are still stochastic, and follow the same rule as the SI model. However, an additional recovery process takes place and it is deterministic: a node remains infectious for a fixed time $\Delta_i$ before recovering permanently. The resulting factor for node $i$ can still be written as in~\eqref{eq:FactSI}, but with the modified infection probability
    \begin{equation}
        {p}^{\rm inf}_i(s,  \{t_k\}_{k\in\partial_i}) = 1 -\prod_{k\in \partial_i} (1-\lambda_{ki}\mathds{I}[t_k < s\leq t_k+\Delta_k])
    \end{equation}
    
    We can recover the SI model by fixing $\Delta_i > T\;\ \forall
    \,i$, such that all nodes remain infectious until the end. 
    
\end{itemize}
Throughout, we assume the form of the spreading kernel in~\eqref{eq:Spread1} is known, as well as all parameters $\vect{\Theta}$ of the spreading model. In the SI model, these include the graph $G$ and all the associated infectivity parameters $\lambda_{ij}\;\forall\,(i,j)\in E$, while for the dSIR model we also have the recovery delays $\Delta_i\;\forall\,i\in V$. Moreover, we will focus on studying inference on graphs generated through random sparse ensembles, such as Erdos-Renyi~(ER) and Random Regular~(RR) graphs.

\subsection{Neural-network priors on the spreading sources}
Spreading models often assume that the nodes that originate the spreading, i.e.\ the sources, are single~\cite{pinto2012locating,shah2010detecting,lokhov2014inferring} or multiple~\cite{luo2013identifying,altarelli2014bayesian} uniformly random nodes. Realistically, the source prior is often non-trivial and can be informed by known node-wise covariates $F\in\mathbb{R}^{N\times M}$, namely each node $i\in V$ has $M$ features, contained in the vector $\vect{F_i}\in\mathbb{R}^M$, representing some contextual information about each node. For instance, in epidemiology, covariates like age or mobility can determine a node's propensity to act as an initial source. Such features can be incorporated to improve inference~\cite{fan2015bayesianmodelsheterogeneouspersonalized}. This implies the initial state is a function of its covariates, $x_i^0 = f(\vect{F_i})$. In the SI and dSIR models described before, this variable is binary: node $i$ can either be a source of the spreading, and we conventionally assign $x_i^0 = -1$, or it is susceptible at the start, and in this case $x_i^0 = +1$. 

Since generic functions can be parametrised using a  multi-layer neural network, it is reasonable to consider the covariates $\vect{F_i}$ to be an input and the initial states $\vect{x_i^0}$ to be an output of a multi-layer fully-connected neural network:
\begin{equation}
    x_i^0 = \varphi^{(L)}\left(W^{(L)}\varphi^{(L-1)}\left(W^{(L-1)}\dots\varphi^{(1)}\left(W^{(1)}\vect{F_i}\right)\dots\right)\right)
\end{equation}
where $\varphi^{(l)}(\cdot)$ are component-wise non-linearities and $\{W^{(l)}\}_{l=1}^L$ are the network weights. We call this the Neural Sources Spreading (NSS) model.

In order to obtain a model where the information contained in the prior can be quantified precisely, we will consider in what follows a single-layer network with random weights $\vect{u}\in\mathbb{R}^M$, following a known, component-wise prior $u_{a} \sim P_u$, $\forall\,a\in\{1,\dots,M\}$. The initial state is then sampled according to
\begin{equation}\label{eq:Perceptron1}
    x_i^0 \sim P^{\rm out}\giventhat{x_i^0}{\vect{F_i},\vect{u}} \equiv \delta\left(x_i^0 - \varphi\left(\sum_{a}F_{ia}u_{a}\right)\right)\,.
\end{equation}
Here, $\delta(\cdot)$ is the Dirac delta. For modelling purposes we also choose $F\in\mathbb{R}^{N\times M}$ as a matrix of i.i.d. Gaussian components $F_{ia}\sim \mathcal{N}(0,1/M)$. We will consider the two cases of Gaussian and Rademacher weights $\vect{u}$. We set $\varphi(x) = \text{sign}(x - \kappa)$, yielding the perceptron model~\cite{gardner1988space,gardner1988optimal,krauth1989storage}, where the threshold $\kappa$ controls the source density. We call this specific version the Perceptron Sources Spreading (PSS) model.

In order to be in an analytically tractable setting, we study the limit where $N, M \to \infty$ with a finite constant ratio $\alpha = N/M = \mathcal{O}(1)$. In this regime, $\alpha$ acts as a signal-to-noise ratio controlling correlations between initial states. For large $\alpha$ (small $M$), the $x_i^0$ are strongly correlated via $\vect{u}$. As $\alpha \to 0$, the terms $\sum_{a=1}^MF_{ia}u_{a}$ become independent Gaussians, and we recover a uniform prior $P(x_i^0 = I) = \delta$ for all $i$, with $\delta = \frac{1}{2} \left(1+\text{erf}\left(\frac{\kappa}{\sqrt{2}}\right)\right)$.

\section{\label{sec:BI}Bayesian inference framework}
In this section, we define in detail the inference problem. 
We assume we are given the covariates $F \in \mathbb R^{N\times M}$, the graph structure $G$ and the edge-wise transmission probabilities $\lambda_{ij}$. We also know the constant $\kappa$ from~\eqref{eq:Perceptron1}, and all $\Delta_i$ for the dSIR model. We are not given the vector $\vect{u}$, and consequently the sources are also unknown, as well as the transition times ${\bf t}_i$. The task is to recover the initial state ${\bf x}^0$ and the transition times ${\bf t}_i$ based on additional partial observation of the state of each node $\{\mathcal{O}_i\}_{i=1}^N$. 
We assume the availability of node-wise observations $\bm{\mathcal{O}} = \{\mathcal{O}_i\}_{i=1}^N$,
\begin{equation}
    P\giventhat{\vect{\mathcal{O}}}{\{\vect{t_i}\}, \vect{x^0},\vect{\Theta}} = \prod_{i=1}^N P\giventhat{\mathcal{O}_i}{\vect{t_i},x_i^0,\vect{\Theta}}
\end{equation}
where $\{\vect{t_i}\}$ is the set of all transition times. This implies that the observation $\mathcal{O}_i$ is conditionally independent of other nodes' transition times given $\vect{t_i}$.
In the following, we will focus on two classes of observations: the first are \textit{sensors}~\cite{pinto2012locating,spinelli2017general}, in which we choose randomly a fraction $\rho$ of the nodes, and we observe their full time trajectory, such that an observation on node $i$ acts as a delta function on the transition time $t_i$. The second framework, known as \textit{snapshots}~\cite{lokhov2014inferring,altarelli2014patient}, is when the entire system state $\vect{x^{t=T_{\rm obs}}}$ is observed at a single time $T_{\rm obs}$, and thus an observation on node $i$ does not fix $t_i$, but gives either a lower bound $t_i\geq T_{\rm obs}$ or an upper bound $t_i < T_{\rm obs}$.

Using Bayes' rule, the posterior distribution of the model becomes:
\begin{align}
   P\giventhat{\{\vect{t_i}\}, \vect{x^0}, \vect{u} }{\vect{\mathcal{O}},F,\vect{\Theta}} = & \frac{1}{P\giventhat{\vect{\mathcal{O}}}{ F,\vect{\Theta}}}P\giventhat{\vect{\mathcal{O}}}{\{\vect{t_i}\}, \vect{x^0}, \vect{u} ,F,\vect{\Theta}}P\giventhat{\{\vect{t_i}\}, \vect{x^0}, \vect{u}}{ F,\vect{\Theta}}\\
   =&\frac{1}{P\giventhat{\vect{\mathcal{O}}}{ F,\vect{\Theta}}}P\giventhat{\vect{\mathcal{O}}}{\{\vect{t_i}\}, \vect{x^0}, \vect{\Theta}}P\giventhat{\{\vect{t_i}\}}{  \vect{x^0}, \vect{\Theta}}P\giventhat{  \vect{x^0}}{ \vect{u}, F}P(\vect{u})\\
   =&\frac{1}{Z(\vect{\mathcal{O}},F,\vect{\Theta})}\prod_{i=1}^N \widetilde{\Psi}_i(\vect{t_i}, \{\vect{t_j}\}_{j\in\partial_i},x_i^0,\mathcal{O}_i, \vect{\Theta}) \Psi_i^{\rm out}(x_i^0, \vect{u}, \vect{F_i})\prod_{a=1}^M\psi_{a}(u_{a})\,,\label{eq:post_eq}
\end{align}
where
$\psi_{a}(u_{a}) = P^u(u_{a})$ is the prior distribution of the weights, $\Psi_i^{\rm out}(x_i^0, \vect{u}, \vect{F_i}) = P^{\rm out}\giventhat{x_i^0}{\vect{F_i},\vect{u}}$ the output distribution of the initial epidemic state of the nodes given the external context, and $\widetilde{\Psi}_i(\vect{t_i}, \{\vect{t_j}\}_{j\in\partial_i},x_i^0,\mathcal{O}_i, \vect{\Theta}) \equiv \Psi_i(\vect{t_i}, \{\vect{t_j}\}_{j\in\partial_i}, \vect{\Theta})P\giventhat{\mathcal{O}_i}{\vect{t_i},x_i^0,\vect{\Theta}}$, which contains the information regarding the epidemic and the observations provided. Finally, 
\begin{align}
Z\left(\vect{\mathcal{O}},F,\vect{\Theta}\right) &\equiv 
\int \left(\prod_{a=1}^M\dd{u}_{a}\psi_{a}(u_{a})\right)\sum_{\{\vect{t_i},x_i^0\}_{i=1}^N} \prod_{i=1}^N \widetilde{\Psi}_i(\vect{t_i}, \{\vect{t_j}\}_{j\in\partial_i},x_i^0,\mathcal{O}_i, \vect{\Theta})\Psi_i^{\rm out}(x_i^0, \vect{u}, \vect{F_i})
\end{align}
is the partition function, and we define 
\begin{equation}\label{eq:FreeEntZ}
\Phi\left(\vect{\mathcal{O}},F,\vect{\Theta}\right) = \frac{1}{N}\log Z\left(\vect{\mathcal{O}},F,\vect{\Theta}\right)
\end{equation}
the associated free entropy.

\subsection{\label{sec:OptInf}Sources retrieval and optimal inference}
A Bayesian approach unifies various inference tasks, as they all reduce to estimating marginals of initial states $\{x_i^0\}$ or transition times $\{\vect{t_i}\}$. For concreteness, we focus on retrieving the spreading sources. An analysis of retrieving the full trajectory, which shows similar phenomenology, is in Appendix~\ref{sec:MSE}.

To quantify source recovery performance, since the initial states $x_i^0$ are binary (susceptible or source), a natural metric is the overlap between the ground-truth state $\vect{x^{*,0}}$ and an estimator $\vect{\hat{x}^{0}}$. The overlap is defined as:
\begin{equation}
    \rm{O}(\vect{x},\vect{y}) \equiv \frac{1}{N}\sum_{i=1}^N\delta_{x_i,y_i}\,,
\end{equation}
where $\vect{x},\vect{y}$ are two generic vectors and $\delta_{x_i,y_i}$ is the Kronecker delta.

The Bayes-optimal strategy uses both observations and features to compute the posterior $P(\vect{x^0}\mid\vect{\mathcal{O}},F)$ and leads to an estimator $\vect{\hat{x}^0}$ that maximises the mean overlap with the true state:
\begin{equation}
    \rm{MO}(\vect{\hat{x}^0}) \equiv \mathbb{E}_{\vect{x^0}\mid\vect{\mathcal{O}},F}\left[\rm{O}(\vect{\hat{x}^0},\vect{x^0})\right] = \frac{1}{N}\sum_{\vect{x^0}}P(\vect{x^0}\mid\vect{\mathcal{O}},F)\sum_{i=1}^N \delta_{\hat{x}_i^0,x_i^0}\,.
\end{equation}
It is known~\cite{zdeborova2016statistical} that this quantity is maximised by the estimator
\begin{equation}\label{eq:MMO}
    \hat{x}_i^{0,\rm{MMO}} = \underset{x_i^0}{\operatorname{argmax}}\,\mu(x_i^0\mid \vect{\mathcal{O}},F)\quad \forall\,i\in[1:N]
\end{equation}
where $\mu(x_i^0\mid \vect{\mathcal{O}},F)$ is the posterior marginal for node $i$. This is the Maximum Mean Overlap (MMO) estimator. We analyse its performance via the overlap ${\rm O}_{t=0}\equiv \rm{O}( \vect{\hat{x}^{0,\rm{MMO}}}, \vect{x^{*,0}})$ and mean overlap ${\rm MO}_{t=0} \equiv {\rm MO}(\vect{\hat{x}^{0,{\rm MMO}}})$.

In a Bayes-optimal setting, the Nishimori conditions have to hold~\cite{nishimori2001statistical,zdeborova2016statistical}. Since we will resort to approximations, verifying the validity of the Nishimori conditions is a useful self-consistency check. 
In a Bayes-optimal setting, indeed, the estimated prior and likelihood match the true distributions, making samples from the posterior statistically indistinguishable from the ground truth. This implies that the expected overlap ${\rm O}_{t=0}$ and the expected mean overlap ${\rm MO}_{t=0}$ must coincide in expectation. The expected overlap over the disorder is:
\begin{equation}\label{eq:Nishi1}
    \mathbb{E}_{\vect{x^{*,0}},\vect{\mathcal{O}},F}\left[\rm{O}( \vect{\hat{x}^{0,\rm{MMO}}}(\vect{\mathcal{O}},F), \vect{x^{*,0}})\right] = \mathbb{E}_{F}\Big[\sum_{\vect{x^{*,0}},\vect{\mathcal{O}}}\rm{O}( \vect{\hat{x}^{0,\rm{MMO}}}(\vect{\mathcal{O}},F), \vect{x^{*,0}}) P^*(\vect{x^{*,0}}) P^*\giventhat{\vect{\mathcal{O}}}{\vect{x^{*,0}}}\Big]
\end{equation}
where $P^*$ denotes the ground-truth distributions. Similarly, the disorder-averaged mean overlap is:
\begin{align}
    \mathbb{E}_{\vect{\mathcal{O}},F}\left[\rm{MO}( \vect{\hat{x}^{0,\rm{MMO}}}(\vect{\mathcal{O}},F))\right] &= \mathbb{E}_{\vect{\mathcal{O}},F}\Big[\sum_{\vect{x^{0}}}\rm{O}( \vect{\hat{x}^{0,\rm{MMO}}}(\vect{\mathcal{O}},F),\vect{x^{0}})  P\giventhat{\vect{x^{0}}}{\vect{\mathcal{O}},F}\Big] = \\
    &= \mathbb{E}_{F}\Big[\sum_{\vect{x^{0}},\vect{\mathcal{O}}}\rm{O}( \vect{\hat{x}^{0,\rm{MMO}}}(\vect{\mathcal{O}},F), \vect{x^{0}}) P\giventhat{\vect{\mathcal{O}}}{\vect{x^{0}}} P(\vect{x^{0}})\Big]\label{eq:Nishi2}
\end{align}
Bayes-optimality implies $P(\cdot) = P^*(\cdot)$, so~\eqref{eq:Nishi1} and~\eqref{eq:Nishi2} coincide. Substituting the MMO estimator yields:
\begin{equation}\label{eq:OV_mu}
    \mathbb{E}[{\rm O}_{t=0}] = \frac{1}{N}\sum_{i=1}^N\mathbb{E}[\mathbb{I}[\argmax_{x_i^0} \mu\giventhat{x_i^0}{\vect{\mathcal{O}},F} = x_i^{*,0}]]
\end{equation}
and 
\begin{equation}\label{eq:MO_mu}
    \mathbb{E}[{\rm MO}_{t=0}] = \frac{1}{N}\sum_{i=1}^N\mathbb{E}[\max_{x_i^0} \mu\giventhat{x_i^0}{\vect{\mathcal{O}},F} ]\,,
\end{equation}
where $\mathbb{I}(\cdot)$ is the indicator function. In our numerical experiments, we utilise the equivalence of the empirical estimates in~\eqref{eq:OV_mu} and~\eqref{eq:MO_mu} to assess the consistency of the approximations made in the algorithm with the expected behaviour of the Bayes-optimal estimator.

\section{\label{sec:BPAMP}The algorithm}
In this section we present the inference algorithm we are going to use to estimate the posterior marginals of the problem. This is derived using the cavity method from statistical physics~\cite{mezard2009information} and combines approximate message passing (AMP) for Generalized Linear Models~\cite{barbier2019optimal,zdeborova2016statistical} with belief propagation (BP) for spreading models~\cite{altarelli2014patient,ghio2023bayes}. 
This approach of merging dense and sparse graphical models has been successfully applied in related contexts~\cite{manoel2017multi,gabrie2018entropy,duranthon2023neural,duranthon2024optimal}.

\begin{figure}
    \centering
    \includegraphics[width=0.7\linewidth]{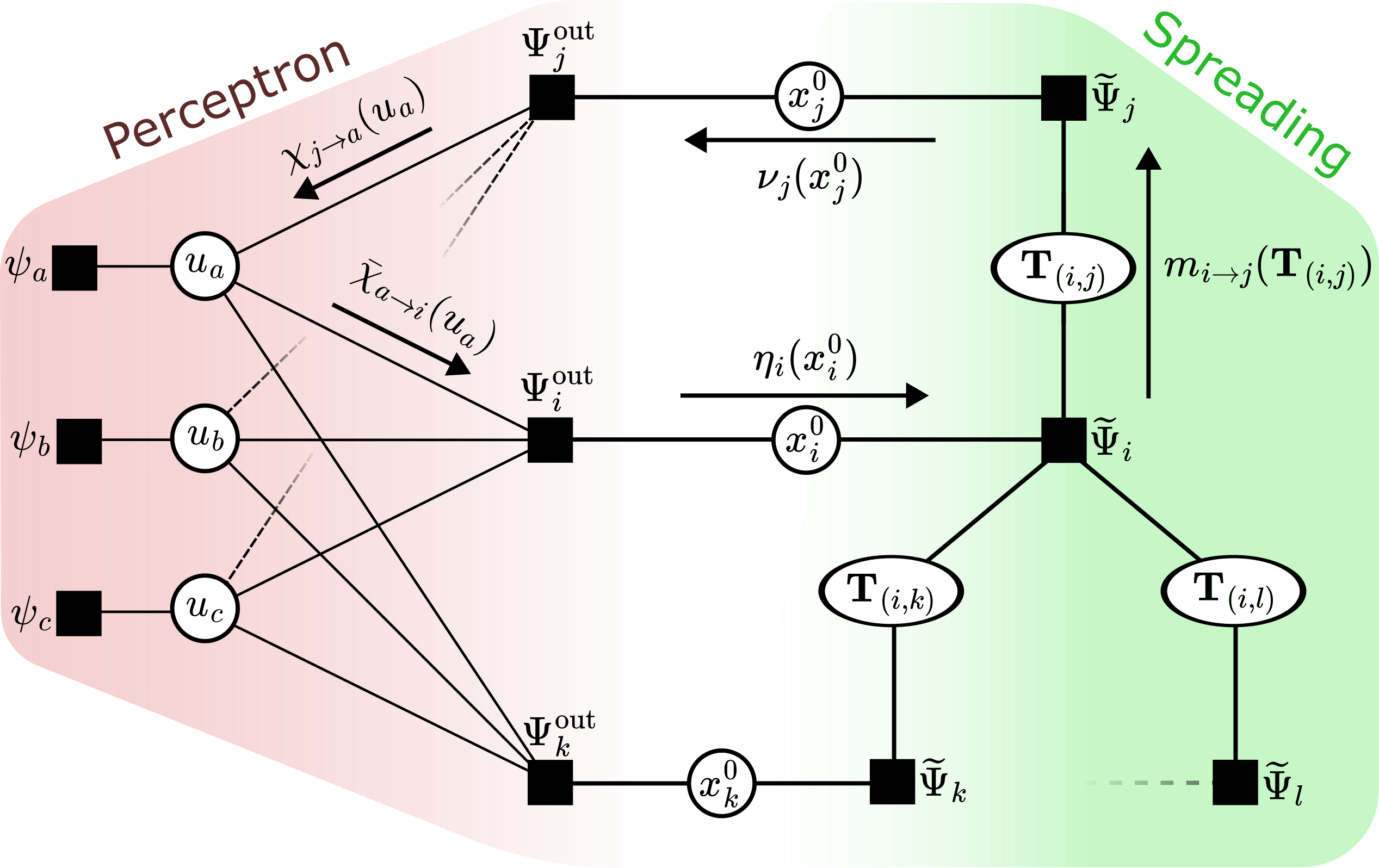}
    \caption{\textbf{Factor Graph.} Representation of the posterior distribution for the Neural Sources Spreading model with a one-layer perceptron prior as a factor graph, with the associated belief propagation messages.}\label{fig:FactorG_GLM}
\end{figure}

The factor graph for the posterior measure in~\eqref{eq:post_eq} is shown in Fig.~\ref{fig:FactorG_GLM}. Standard cavity arguments~\cite{mezard2009information} yield the BP equations:
\begin{equation}
\label{BP-AMP}
\begin{cases}
    m_{i\rightarrow j}(t_i,t_j) &= \frac{1}{Z_{i\rightarrow j}}\sum_{x_i^0}\eta_i(x_i^0)\sum_{\vect{t_{k\in\partial i\backslash j} }}\widetilde{\Psi}_i(t_i,\vect{t_{k\in\partial i}} ,\mathcal{O}_i, x_i^0) \prod_{k\in\partial_i\backslash j} m_{k\rightarrow i}(t_k,t_i) \\
    \nu_i(x_i^0) &= \frac{1}{Z_i^{\nu}}\sum_{t_i,\vect{t_{k\in\partial i}}}\widetilde{\Psi}_i(t_i,\vect{t_{k\in\partial i}} ,\mathcal{O}_i, x_i^0)\prod_{k\in\partial i}m_{k\rightarrow i}(t_k,t_i) \\
    \eta_i(x_i^0) &= \frac{1}{Z_i^{\eta}}\prod_{a}\left(\int {\rm d}u_{a} \bar{\chi}_{a\rightarrow i}(u_a)\right)\Psi_i^{\rm out}(x_i^0,\vect{u},\vect{F_i}) \\
    \chi_{i\rightarrow a}(u_{a}) &= \frac{1}{Z_{i\rightarrow a}}\sum_{x_i^0}\nu_i(x_i^0)\int\left(\prod_{b\neq a}{\rm d}u_{b}\bar{\chi}_{b\rightarrow j}(u_{b})\right)\Psi_i^{\rm out}(x_i^0,\vect{u},\vect{F_i}) \\
     \bar{\chi}_{a  \rightarrow i}(u_{a}) &= \frac{1}{\bar{Z}_{a\rightarrow i}}\psi_{a}(u_{a})\prod_{j\neq i}\chi_{j\rightarrow a}(u_{a})
\end{cases}
\end{equation}
These equations allow to compute the probability marginals exactly when the factor graph associated to the posterior is acyclic~\cite{yedidia2003understanding}. In our setting, we will consider spreading graphs which are locally tree-like, i.e.\ with loops of typical length scaling as $\log N$, for which BP is expected to estimate the marginals with an error which goes to zero as $N$ goes to $\infty$, if the replica symmetric (RS) assumption holds. We will refer to this property as the asymptotic optimality of the algorithm. Recently, some evidence was shown~\cite{braunstein2023statistical,braunstein2025evidence} towards the existence of a replica symmetry broken (RSB) phase in spreading models for very low values of the fraction of sources, where these assumptions break down and the BP equations stop converging. Here, we focus on cases where the density of sources is high enough that these issues are not observed, thus remaining in the regime where the RS assumption is valid.

~\eqref{BP-AMP} are generally intractable due to high-dimensional integrals. They simplify by applying the central limit theorem and projecting messages onto their first two moments, a standard procedure~\cite{zdeborova2016statistical}. These simplifications yield key denoising functions, for which we postpone the detailed derivation in Appendix~\ref{App:GLM_deriv}.

The output denoising function is:
\begin{equation}
g_o(\omega_i,\nu_i,V_i) = \frac{\displaystyle \int \mathrm{d}z\,\sum_{x} \nu_i(x)\,P_{\rm out}(x\mid z)\,(z-\omega_i)\,e^{-\frac{(z-\omega_i)^2}{2V_i}}}{\displaystyle V_i\int \mathrm{d}z\,\sum_{x} \nu_i(x)\,P_{\rm out}(x\mid z)\,e^{-\frac{(z-\omega_i)^2}{2V_i}}} = \dfrac{\,2(2\nu_i-1)\,\exp\Bigl(-\dfrac{(\omega_i-\kappa)^2}{2V_i}\Bigr)}
{\sqrt{2\pi V_i}\left(1+(2\nu_i-1)\,\operatorname{erf}\Bigl(\dfrac{\omega_i-\kappa}{\sqrt{2V_i}}\Bigr)\right)}\,,
\end{equation}
where the last step uses $P_{\rm out}(x\mid z) = \delta[x-\text{sign}(z-\kappa)]$. The input denoising functions are:
\begin{equation}
f_a(A,B) = \frac{\displaystyle \int \mathrm{d}u\,P^u(u)\,u\,\exp\!\Bigl[-\frac{A}{2}\,u^2 + B\, u\Bigr]}{\displaystyle \int \mathrm{d}u\,P^u(u)\,\exp\!\Bigl[-\frac{A}{2}\,u^2 + B\, u\Bigr]},\qquad f_v(A,B) = \frac{\partial}{\partial B} f_a(A,B).
\end{equation}
For a Gaussian prior, $f_a(A,B) =\frac{B}{A+1}$ and $f_v(A,B) = \frac{1}{A+1}$. For a Rademacher prior, $f_a(A,B) =\tanh(B)$ and $f_v(A,B) = 1 - \tanh^2(B)$. The final algorithm is presented in Algorithm~\ref{Algo_GLM}, and it allows getting directly the cavity estimation of the marginals for the variables of the problem.
For the sake of computing the overlaps defined in Sec.~\ref{sec:OptInf}, all we need are the values at convergence of the BP-AMP messages $\eta_i(x_i^0)$ and $\nu_i(x_i^0)$ for every node. Then, the estimation of the marginal 
$\mu\giventhat{x_i^0}{\vect{\mathcal{O}},F}$ will be given by
\begin{equation}
    \chi_i(x_i^0) \equiv \frac{\eta_i(x_i^0)\nu_i(x_i^0)}{\sum_{x_i^0}\eta_i(x_i^0)\nu_i(x_i^0)}\,,
\end{equation}
In turn, we can use this quantity in~\eqref{eq:OV_mu} and~\eqref{eq:MO_mu} to get an estimation of the average overlap and mean overlap.

\begin{algorithm}[t!]
\caption{\label{Algo_GLM}BP-AMP algorithm for the PSS model.}
    \SetKwInOut{Input}{Input}
    \SetKwInOut{Init}{Initialisation}
    \SetKwInOut{Output}{Output}
    \Input{Features $\vect{F}$, Graph $G(N,M)$, Observations $\mathcal{O}$, Functions $g_o$, $f_a$, $f_v$}
    \Init{ $\vect{g_o^{(0)}} = \vect{0}$, $\vect{a^{(0)}}$, $\vect{v^{(0)}}$, $\vect{\nu^{(0)}}$, $\{\vect{m_{i\rightarrow j}^{(0)}}\}_{(i,j)\in E}$}
    \Output{$\vect{\hat{x}^0}$}
    \SetKwBlock{Begin}{Repeat until convergence}{end}
    \Begin{
        $V^{(t+1)} = \frac{1}{M}\sum_{a}v_{a}^{(t)}$\;
        \For{$i = 1,\,\dots,\,N$}{
            $\nu_i^{(t+1)}(x_i^0) = \sum_{t_i,\vect{t_{k\in\partial i}}}\widetilde{\psi}_i(t_i,\vect{t_{k\in\partial i}} ,\mathcal{O}_i, x_i^0)\prod_{k\in\partial i}m^{(t)}_{k\rightarrow i}(t_k,t_i)$\;
            \CommentSty{Normalize $\nu_i$}\;
        }
        
        \For{$i = 1,\,\dots,\,N$}{
            $\omega_i^{(t+1)} = \sum_{a}F_{ia}a_{a}^{(t)}-V^{(t+1)}g_{o,i}^{(t)}$\;
            $g_{o,i}^{(t+1)} = g_o(\omega_i^{(t+1)},\nu_i^{(t+1)},V^{(t+1)})$\;
            $\eta_i^{(t+1)}(x_i^0) = \int {\rm d}z P_{\rm out}(x_i^0\mid z)e^{-\frac{(z-\omega_i^{(t+1)})^2}{2V^{(t+1)}}}$\;
            \CommentSty{Normalize $\eta_i^{(t+1)}$}\;
            $\chi_i^+ = \frac{\eta_i(+1)\nu_i(+1)}{\eta_i(+1)\nu_i(+1) + \eta_i(-1)\nu_i(-1)}$\;
            $\hat{x}_i^{0,(t+1)} = 2\chi_i^+-1$\;
        }
        $A^{(t+1)} = \frac{1}{M} \sum_i\left(g_{0,i}^{(t+1)}\right)^2$\;
        
        \For{$a = 1,\,\dots,\,M$}{
            $B_{a}^{(t+1)} = \sum_iF_{a i}g_{o,i}^{(t+1)}+a_{a}^{(t)}A^{(t+1)}$\;
            $a_{a}^{(t+1)} = f_a(A^{(t+1)},B_{a}^{(t+1)})$\;
            $v_{a}^{(t+1)} = f_v(A^{(t+1)},B_{a}^{(t+1)})$\;
        }
        
        \For{$(i,j) \in E$}{
            $m^{(t+1)}_{i\rightarrow j}(t_i,t_j) = \sum_{x_i^0}\eta_i^{(t+1)}(x_i^0)\sum_{t_i,\vect{t_{k\in\partial i\backslash j} }}\widetilde{\psi}_i(t_i,\vect{t_{k\in\partial i}} ,\mathcal{O}_i, x_i^0) \prod_{k\in\partial_i\backslash j} m^{(t)}_{k\rightarrow i}(t_k,t_i)$\;
            \CommentSty{Normalize $m_{i\rightarrow j}$}\;
        }
        
    }
\end{algorithm}

\subsection{Free entropy}
We have defined the free energy of the problem in~\eqref{eq:FreeEntZ} as the logarithm of the partition function, which in general is a quantity that is too expensive to compute by brute-force. The algorithm just defined allows us to estimate the large size limit of the free entropy directly from the values of the BP-AMP messages at convergence. A standard replica-symmetric cavity computation (see Appendix~\ref{app:FreeEnt}) gives the final expression:
\begin{align}\label{eq:PhiRS}
    \phi_{\rm RS} = \frac{1}{N}&\sum_{i=1}^N \log Z_i^{\nu}  - \frac{1}{N} \sum_{(i,j)\in E} \phi_{(i,j)}^{\rm spread}  + \sum_{a=1}^M \log \int \mathrm{d}u_a\,P^u(u_a)\,\exp\Bigl[-\frac{A}{2}u_a^2 + B_a u_a\Bigr]\\
    &+\frac{1}{N}\sum_{i=1}^N \phi_i^{\rm out}+\sum_{a=1}^M \Bigl[\frac{A}{2}(a_a^2+v_a) - B_a\,a_a\Bigr] + \sum_{i=1}^N \frac{\Bigl(\omega_i - \sum_{a=1}^M F_{ia}a_a \Bigr)^2}{2V_i}\,,
\end{align}
where
\begin{equation}
    \phi_{(i,j)}^{\rm spread} = \log \sum_{t_i,t_j} m_{i\to j}(t_i,t_j)\, m_{j\to i}(t_j,t_i)\,,
\end{equation}
and
\begin{align}
    \phi_i^{\rm out} &= \log\left(\sum_{x_i^0}\nu_i(x_i^0)\int\left(\prod_{b}{\rm d}u_{b}\right)\Psi_i^{\rm out}(x_i^0,\vect{u},\vect{F_i})\prod_{b}\bar{\chi}_{b\rightarrow i}(u_{b})\right) \\
    &=\log\left(\sum_{x_i^0}\nu_i(x_i^0)\eta_i(x_i^0)\right) - \log Z_i^{\eta}\,.
\end{align}
Algorithm~\ref{Algo_GLM} typically has a single fixed point corresponding to the free entropy maximum. However, first-order phase transitions can lead to multiple fixed points. Comparing their free entropy values, and selecting the fixed point with larger free entropy, characterises the Bayes-optimal estimator, a point we revisit at the end of Sec.~\ref{sec:Rade}.

\section{\label{sec:Gauss}Gaussian weights perceptron prior}
We analyse Algorithm~\ref{Algo_GLM}'s performance for source retrieval with Gaussian weights ($P^u(u_{a}) = e^{-u_{a}^2/2}/\sqrt{2\pi}$). 
Our experimental setup is as follows:
\begin{itemize}
    \item \textit{Graph}: We use Random Regular (RR) graphs with degree $d=3$ and homogeneous weights $\lambda_{ij}^t = \lambda$. On these locally tree-like graphs, BP is expected to be asymptotically optimal in the absence of first-order phase transitions~\cite{mezard2009information,coja2017information} and of replica symmetry breaking~\cite{braunstein2023statistical,braunstein2025evidence}. Appendix~\ref{App:Ensemble} shows results for other ensembles, suggesting marginal dependence on network topology.
    \item \textit{Observations}: We consider observations of two types:
    \begin{itemize}
        \item \textit{Sensors}~\cite{pinto2012locating,spinelli2017general}: The full time trajectory of a fraction $\rho$ of random nodes is observed.
        \item \textit{Snapshot}~\cite{lokhov2014inferring,altarelli2014patient}: The entire system state $\vect{x^{t=T_{\rm obs}}}$ is observed at a single time $T_{\rm obs}$.
    \end{itemize}
    \item \textit{Spreading dynamics}: The dynamics follow the dSIR model (Section~\ref{sec:Model}) with a fixed recovery time $\Delta_i = \Delta$. We compare the case $\Delta=1$, in which nodes remain infectious only for a time step, with the standard SI model. In both cases, we run the dynamics until there is no more infectious individual in the network, and we define $T$ as the first time step in which this happens.
\end{itemize}

We measure performance using a rescaled overlap for source retrieval:
\begin{equation}\label{eq:resOv}
    \widetilde{\rm O}_{t=0} = \frac{{\rm O}_{t=0}-{\rm O}( \vect{\hat{x}^{0,\rm{RND}}}, \vect{x^{*,0}})}{1-{\rm O}( \vect{\hat{x}^{0,\rm{RND}}}, \vect{x^{*,0}})}\,,
\end{equation}
where $\vect{\hat{x}^{0,\rm{RND}}}$ is a random estimator, equivalent to the MMO estimator in~\eqref{eq:MMO} but without access to observations ($\vect{\mathcal{O}}=\vect{\emptyset}$). This rescaled overlap is non-negative and equals one at perfect recovery.

\begin{figure}
    \centering
    \includegraphics[width=\linewidth]{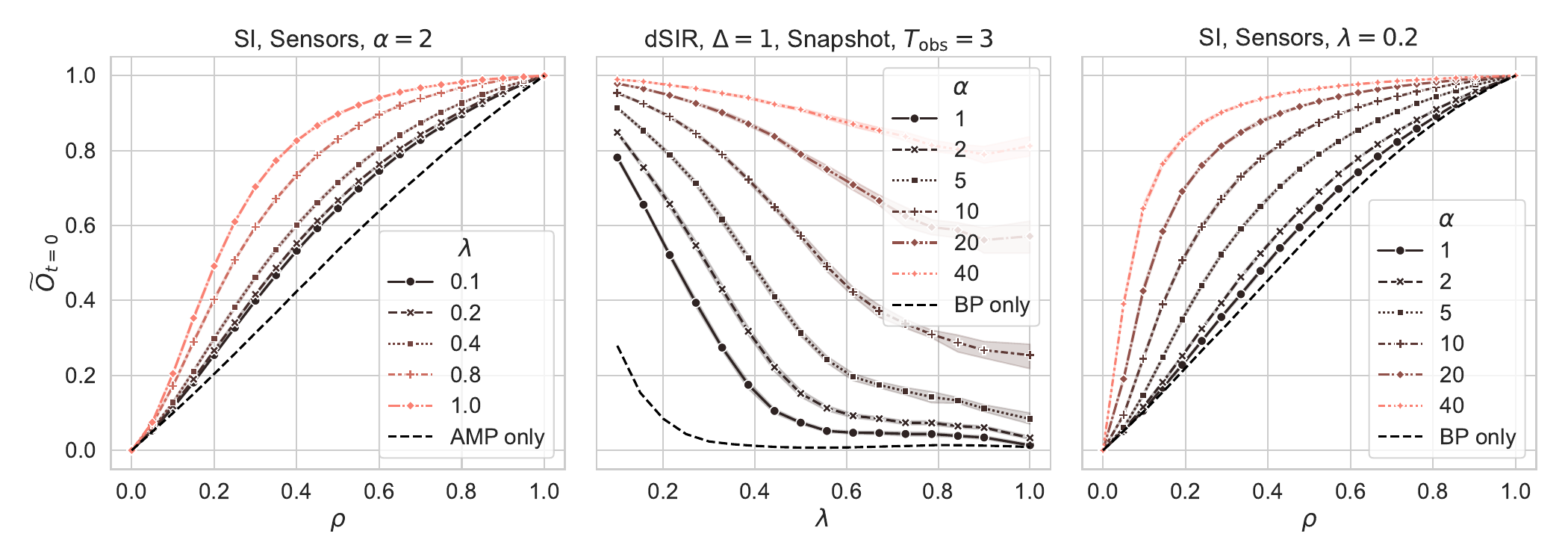}
    \caption{\textbf{Inference performance vs. the transmission rate $\lambda$ and the strength of correlations in the prior $\alpha$.} PSS model, Gaussian weights of the perceptron prior, $\kappa=-1$ ($\delta \approx 0.159$ sources). RRGs ($d=3, N \times M = 1.6\times 10^9$). Average of 20 runs; shading is 99\% confidence interval. 
    In the left and right panels, we take the case of sensor observations, showing the gain in performance when diminishing transmission rate $\lambda$ and when increasing correlation in the prior through $\alpha$, respectively. We also compare them to the baselines we defined in the text, AMP-only and BP-only, respectively. In the central panel, instead, the case of snapshot observations is considered, showing again a considerable increase in performance when using BP-AMP compared to only using BP.
    }\label{fig:PhD_Gauss_lam}
\end{figure}

Fig.~\ref{fig:PhD_Gauss_lam} shows the impact of the infectivity of the epidemic process $\lambda$ and the signal-to-noise ratio $\alpha=N/M$ on the algorithmic performance. Moreover, we compare the performance of the AMP-BP algorithm to two other algorithms:
\begin{itemize}
    \item \textbf{BP-only}: The node covariates $F$ are not disclosed, and we rely only on the Belief Propagation for the inference (this corresponds to the $\alpha\to0$ limit studied in~\cite{ghio2023bayes}).
    \item \textbf{AMP-only}: In the case of sensor observations, we can consider the case in which BP is not run on the graph, but we are only allowed to use the $\rho N$ observations as input variables for the AMP algorithm on the classical perceptron problem (this corresponds to the analysis done in~\cite{barbier2019optimal}).
\end{itemize}

The left panel of Fig.~\ref{fig:PhD_Gauss_lam} shows the rescaled overlap for the SI model with sensor observations versus the sensor fraction $\rho$. From this plot we can evince that lower stochasticity (higher $\lambda$) yields greater performance achieved from BP and equivalently a higher gain compared to the AMP-only case. Even at low $\lambda$, where spreading inference is harder, a significant performance gain is observed.

The central panel of Fig.~\ref{fig:PhD_Gauss_lam} uses snapshot observations at fixed $T_{\mathrm obs}=3$ for dSIR ($\Delta=1$). Here, inference quality depends on the outbreak size, controlled by $\lambda$. For small $\lambda$, the localised outbreak allows the snapshot to pinpoint sources, yielding near-perfect recovery. For large $\lambda$, most nodes are already infected and recovered, so the snapshot provides little topological information, limiting performance, especially for low $\alpha$. Again, we notice an increased performance when using the information from the node covariates ($\alpha>0$) compared to the BP-only case.

Finally, in the right panel we fix $\lambda=0.2$, and we consider sensors in the SI model. By varying the sensors fraction $\rho$, we look at different values of $\alpha$, so that we can compare again to the BP-only performance. We see also in this case that this baseline is recovered in the $\alpha\to0$ limit, while for larger $\alpha$ the inference capability of the algorithm are greatly increased.

\begin{figure}
    \centering
    \includegraphics[width=1\linewidth]{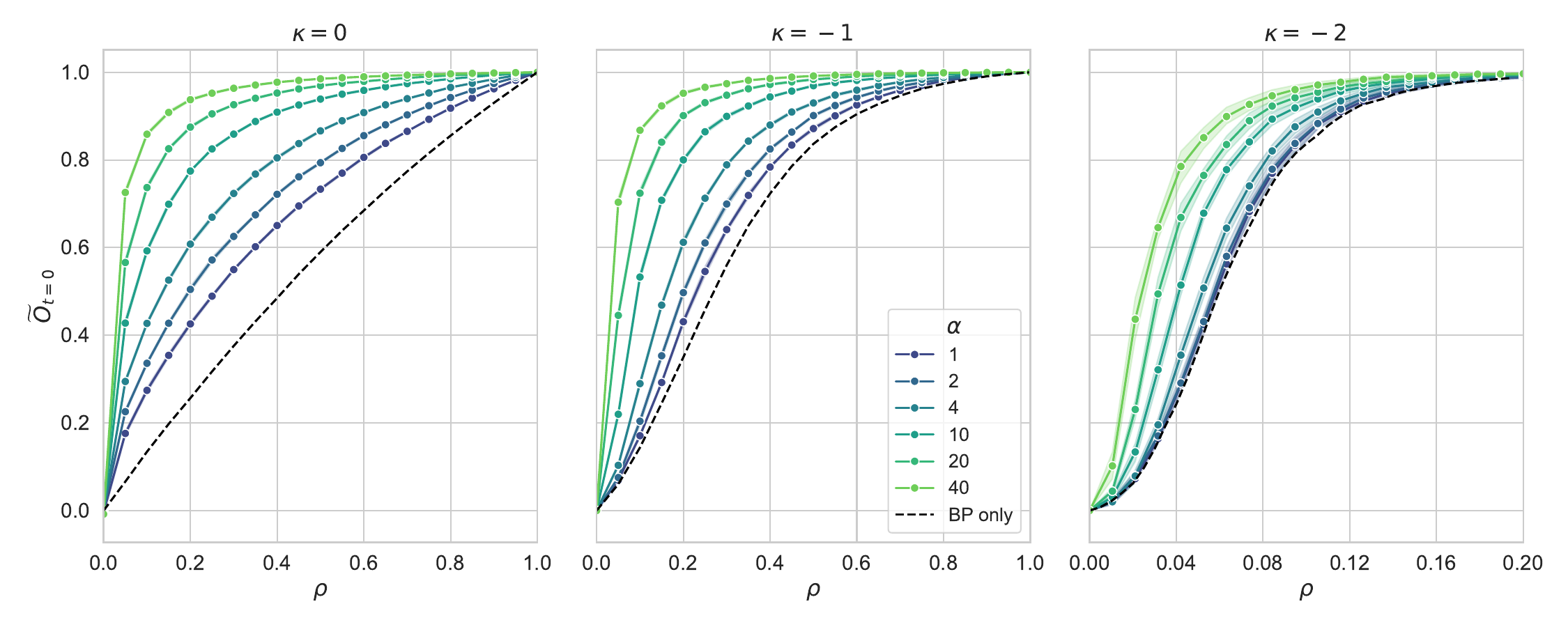}
    \caption{\textbf{Inference performance versus fraction of sensor observations.} 
    PSS model, Gaussian weights of the perceptron, varying correlation in the prior $\alpha$ and density of sources $\kappa$ (source fractions $\delta \approx0.5, 0.159, 0.023$). SI model ($\lambda=1$) with sensor observations. RRGs ($d=3, N=20000$). Rescaled overlap vs. sensor fraction $\rho$. Average of 20 runs; shading is 99\% confidence interval. The dashed lines are the performance of the BP algorithm, ignoring the part from the prior. We can see that BP-AMP improves substantially over this baseline, gaining more as the correlation between the initial states increases.}\label{fig:PhD_Gauss}
\end{figure}

For the SI model without stochasticity ($\lambda=1$), Fig.~\ref{fig:PhD_Gauss} shows performance versus sensor fraction $\rho$ for varying $\alpha$ and $\kappa$. As expected, overlap increases with $\alpha$, which acts as a signal-to-noise ratio. As $\alpha \to 0$, performance approaches that of the BP-only estimator (no neural-network prior). The smoothness of all curves confirms the absence of phase transitions in this setting.

\section{\label{sec:Rade}Binary weights perceptron prior}
We now analyse the case of binary weights of the perceptron prior using a Rademacher prior, i.e., $P^u(u_{a}) = \frac{1}{2}\delta_{u_{a},+1} + \frac{1}{2}\delta_{u_{a},-1}$. This choice reveals a different phenomenology compared to the Gaussian case, and we investigate its dependence on model parameters.

For concreteness, we fix the spreading model to be SI on a Random Regular Graph (RRG) with degree $d=3$. As shown in Appendix~\ref{app:AddPlot}, these choices do not qualitatively affect the main results of this section. We use sensor observations, with the sensor fraction $\rho$ acting as the information control parameter, and assess source retrieval via the overlap parameter in~\eqref{eq:resOv}. Other inference tasks are deferred to Appendix~\ref{app:AddPlot}. 

Before discussing the results, we comment on the Nishimori conditions. While the Gaussian case showed rapid convergence to validity of the Nishimori conditions necessary for Bayes-optimality (Fig.~\ref{fig:Nishi}), the Rademacher case exhibits significant finite-size effects near the perfect recovery transition we show below. These manifest in a small fraction of realisations where the mean overlap jumps to one at the fixed point, while the overlap remains finite. We show in~Fig.~\ref{fig:Nishi_Rade} that this discrepancy diminishes with system size, and we conjecture that it vanishes in the thermodynamic limit. In the plots below, we neglect these few atypical realisations, in order to better reflect the behaviour of the model in typical instances.

\begin{figure}
    \centering
    \includegraphics[width=1\linewidth]{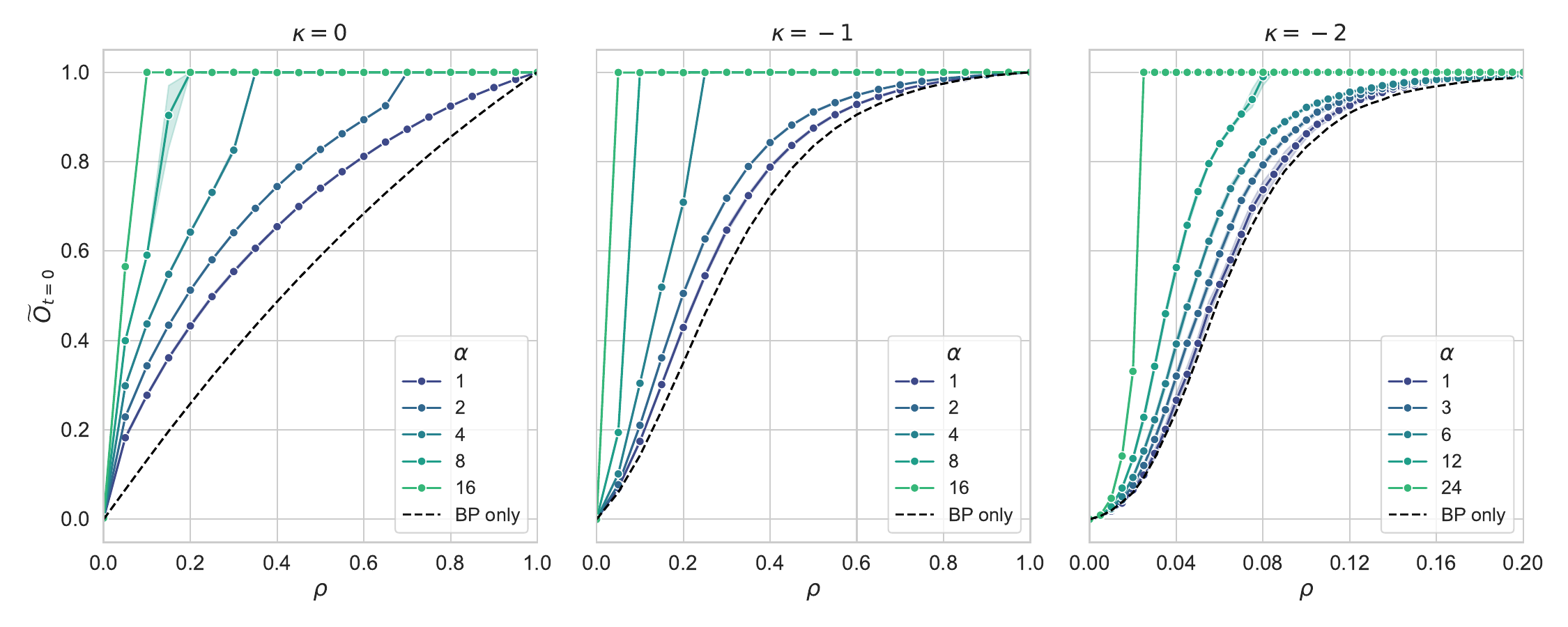}
    \caption{\textbf{Inference performance vs fraction of sensors for perceptron prior with Rademacher weights.} PSS model, varying $\alpha=N/M$ and threshold $\kappa=0,-1,-2$ (average source fractions $\delta \approx0.5, 0.159, 0.023$). Sensor observations for SI model ($\lambda=1$). Simulations on RRGs ($d=3, N \times M = 1.6\times 10^9$). Rescaled overlap in~\eqref{eq:resOv} vs. sensor fraction $\rho$. Each point is an average of 20 runs starting from random initialisation; shaded region is the 99\% confidence interval. We observe a significant gain when using BP-AMP compared to BP alone. Compared to the Gaussian case in~Fig.~\ref{fig:PhD_Gauss}, where the curves remained continuous, here for $\alpha >\alpha_c$, studied in detail later, the algorithm achieves perfect recovery and the overlap jumps to one.}\label{fig:PhD_Rade}
\end{figure}

Fig.~\ref{fig:PhD_Rade} shows the algorithm's performance for the non-stochastic SI model ($\lambda=1$). Each panel plots the rescaled overlap $\widetilde{\mathrm O}_{t=0}$ against sensor density $\rho$ for a fixed $\kappa$ and various ratios $\alpha$. The dashed lines indicate the BP-only baseline. For small $\alpha$, the curves follow the baseline. For intermediate and large $\alpha$, we observe a sharp, discontinuous transition in the overlap, an effect more pronounced for more negative $\kappa$ (sparser sources). This transition leads to perfect recovery ($\widetilde{\mathrm O}_{t=0}=1$) at a critical density $\rho_{\mathrm c}(\alpha,\kappa)$. In the next subsection, we explore the algorithmic implications of this transition.

\subsection{First-order phase transition and hardness}
We now analyse the transition from Fig.~\ref{fig:PhD_Rade} using the free entropy (Sec.~\ref{sec:BPAMP}). To probe the landscape for multiple stable states, we compare two initialisations:
\begin{itemize}
    \item Random: No observations ($\vect{\mathcal{O}} = \emptyset$). BP messages are initialised uniformly. AMP variables are set as $\vect{a} \sim \mathcal{N}(0,1/N)$ and $\vect{v} = \vect{1}$.
    \item Informed: In addition to observations $\vect{\mathcal{O}}$, the true initial state $\vect{x^{*,0}}$ is provided. BP messages are uniform, but AMP variables are set to the ground truth $\vect{a} = \vect{u}$ and $\vect{v} = \vect{1}/\sqrt{N}$.
\end{itemize}
These initialisations are designed to find two key fixed points: a ``perfect recovery'' state and a ``partial recovery'' state. The stability and free entropy of these fixed points characterise the first-order phase transition.

\begin{figure}
    \centering
    \includegraphics[width=1\linewidth]{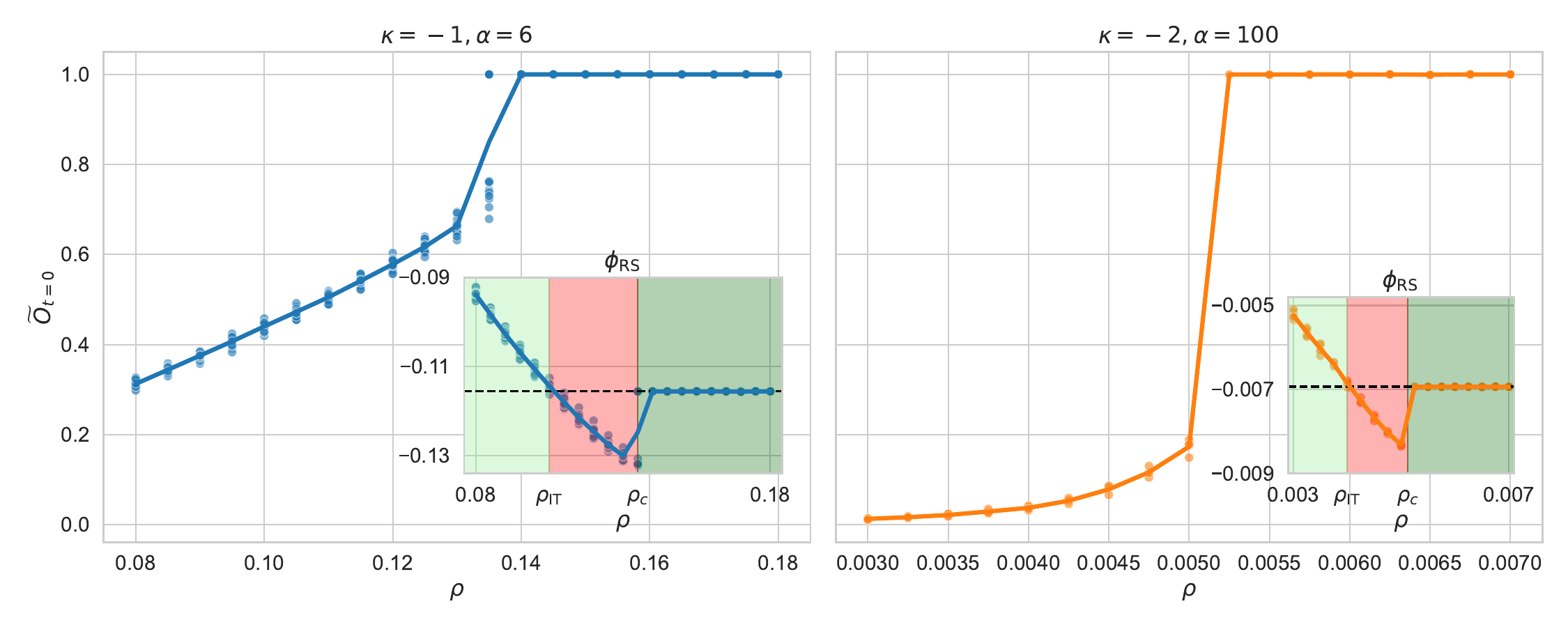}
    \caption{\textbf{Perfect recovery transition and Hard Phase.} PSS model with Rademacher weights. Left: $\kappa=-1$ ($\delta \approx 0.159$), $\alpha=6$. Right: $\kappa=-2$ ($\delta \approx 0.023$), $\alpha=100$. SI model ($\lambda=1$) with sensor observations. RRGs ($d=3, N \times M = 2.5\times 10^9$). In the main plots we show that the rescaled overlap has a jump when changing $\rho$, reaching perfect recovery for a certain $\rho_c$. At the same time, the free entropy (insets) shows the typical behaviour of first order transitions: the onset of the hard phase (shown in red) happens at $\rho_{\rm IT}$, while at $\rho_c$ there is perfect recovery and for all $\rho > \rho_c$ the free energy coincides with the one of the informed fixed point (shown as a dashed black line).}\label{fig:zoom_logZ}
\end{figure}

Using the sensor fraction $\rho$ as the control parameter, we define $\rho_c$ as the spinodal point where the ``partial recovery'' fixed point becomes unstable. For $\rho > \rho_c$, only the ``perfect recovery'' fixed point is stable, and the algorithm finds all sources, even from a random start. The ``perfect recovery'' fixed point is numerically stable for all $\rho \in [0,1]$. Thus, for $\rho < \rho_c$ two scenarios arise based on which fixed point has a higher free entropy: 1) If the ``partial recovery'' fixed point is dominant, the algorithm's performance is Bayes-optimal. 2) If the ``perfect recovery'' fixed point is dominant, the algorithm gets trapped in the metastable ``partial recovery'' state. This defines a computationally hard phase. The threshold between these cases is $\rho_{\rm IT}$ and  if $\rho_{\rm IT} < \rho_c$ the hard phase exists for all $\rho \in [\rho_{\rm IT}, \rho_c]$.

Fig.~\ref{fig:zoom_logZ} illustrates this phenomenon for two $(\kappa, \alpha)$ pairs that exhibit a transition to perfect recovery ($\rho_c < 1$). As shown in Appendix~\ref{app:FreeEnt}, the free entropy of the informed fixed point for $\lambda=1$ is $\phi_\mathrm{info} = -\log 2/\alpha$. By comparing the free entropy from a random start, $\phi_\mathrm{RS}$, to $\phi_\mathrm{info}$, we compute the information-theoretic threshold $\rho_\mathrm{IT}$. The insets show that for these parameters, a finite interval $\rho\in(\rho_\mathrm{IT},\rho_\mathrm{c})$ forms a computationally hard phase: Bayes-optimal recovery is possible, but our algorithm fails to find it.

\begin{figure}
    \centering
    \includegraphics[width=1\linewidth]{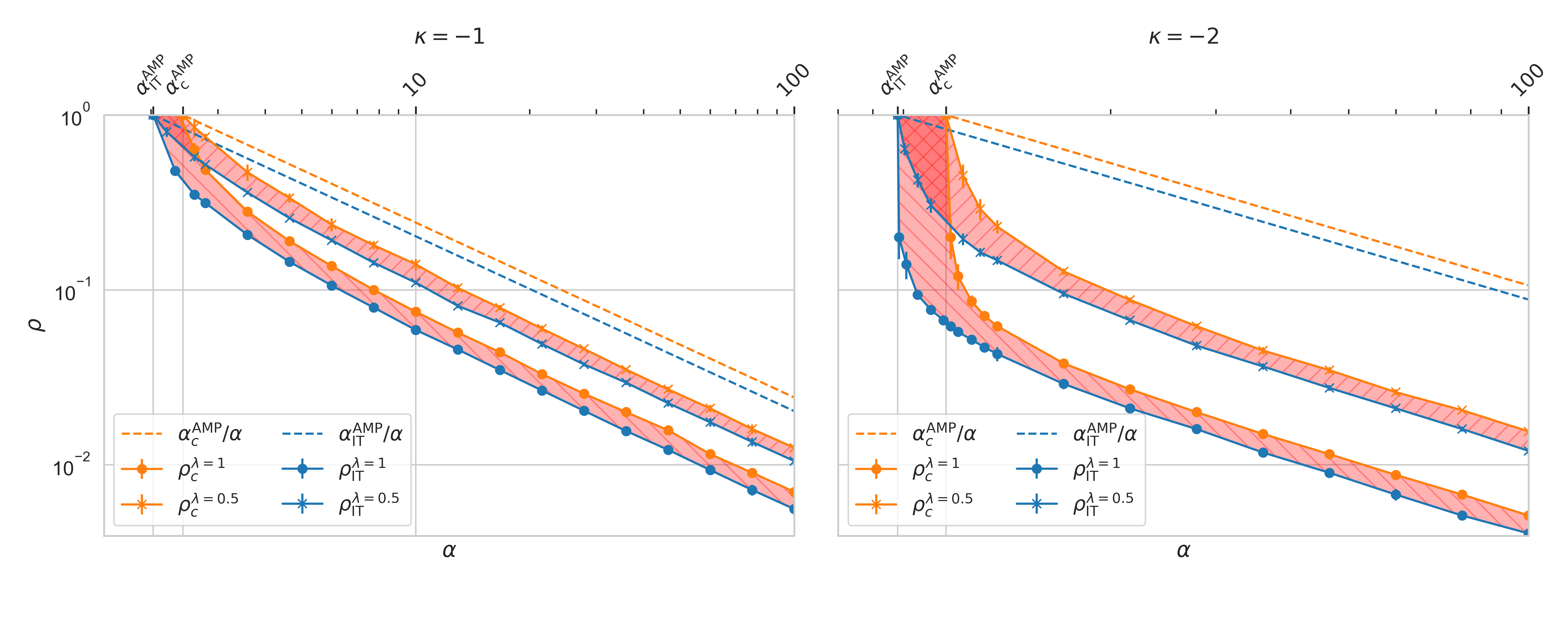}
    \caption{\textbf{First-order transition thresholds.} PSS model with Rademacher weights. Left: $\kappa=-1$ ($\delta \approx 0.159$). Right: $\kappa=-2$ ($\delta \approx 0.023$). SI model ($\lambda\in\{0.5,1\}$) with sensor observations. RRGs ($d=3, N\cdot M \approx 2.5\times10^9$). Points show $\rho_\mathrm{IT}$ and $\rho_\mathrm{c}$ for BP-AMP, i.e.\ the information-theoretic threshold and the spinodal, respectively. Dashed lines are the same thresholds for the AMP-only problem, and we show how both move considerably to smaller values of $\rho$ when introducing the information coming from BP, especially for low values of stochasticity (high values of $\lambda$).}\label{fig:phase_diag}
\end{figure}

Fig.~\ref{fig:phase_diag} shows how the thresholds $\rho_\mathrm{c}$ and $\rho_\mathrm{IT}$ depend on $\alpha$ and $\lambda$. We compare these to the equivalent thresholds for the AMP-only perceptron problem (using $\rho N$ as input dimension), where then $\rho^\mathrm{AMP}_\mathrm{c}(\kappa,\alpha) = \alpha_\mathrm{c}^\mathrm{AMP}(\kappa)/\alpha$ and $\rho^\mathrm{AMP}_\mathrm{IT}(\kappa,\alpha) = \alpha_\mathrm{IT}^\mathrm{AMP}(\kappa)/\alpha$. The values for $\alpha^\mathrm{AMP}_\mathrm{c/IT}$ can be computed from State Evolution~\cite{barbier2019optimal}. For high $\alpha$, our BP-AMP algorithm requires a smaller sensor density for perfect recovery than pure AMP, and interestingly both follow a $\rho\sim1/\alpha$ scaling. The curves converge at $\rho=1$ and $\alpha = \alpha_{\rm c / IT}^{\rm AMP}$, since this corresponds to observing the whole system.

This behaviour defines three distinct regions at fixed $\alpha$, as $\rho$ increases, which we illustrate more in detail in Appendix~\ref{app:PhT}:
\begin{itemize}
    \item $\alpha<\alpha_{\rm IT}^{\rm AMP}$: The algorithm is Bayes-Optimal but never achieves perfect recovery, as it is information-theoretically impossible.
    \item $\alpha_{\rm IT}^{\rm AMP}<\alpha<\alpha_{\rm c}^{\rm AMP}$: A hard phase exists for $\rho > \rho_{\rm IT}^{\lambda}(\kappa)$. Perfect recovery is theoretically possible but algorithmically unreachable, as the spinodal $\rho_c > 1$.
    \item $\alpha>\alpha_{\rm c}^{\rm AMP}$: All three regimes appear: Bayes-optimal for $\rho < \rho_{\rm IT}^{\lambda}(\kappa)$, a hard phase for $\rho_{\rm IT}^{\lambda}(\kappa)< \rho <\rho_{\rm c}^{\lambda}(\kappa)$, and finally perfect recovery for $\rho > \rho_{\rm c}^{\lambda}(\kappa)$.
\end{itemize}

Looking at Fig.~\ref{fig:phase_diag}, we can also see how the spreading parameters affect these phases. A smaller source fraction (more negative $\kappa$) enhances the gain from BP, shifting the hard phase to lower sensor densities. Similarly, increasing stochasticity (smaller $\lambda$) decreases the gap with the AMP-only thresholds.

Finally, we comment on the relevance of these phase transitions to algorithmic hardness. While proving statements about all polynomial-time algorithms is beyond the scope of this work, significant results have established the asymptotic optimality of AMP among large classes of algorithms for various inference tasks~\cite{celentano2020estimation,montanari2024equivalence}. Motivated by this line of research, we conjecture that our BP-AMP algorithm is also asymptotically optimal within these classes. Consequently, the hardness phenomena identified here likely represent fundamental computational hurdles.

\newpage
\section{Conclusions}
In this work, we introduced a model for local spreading on graphs where sources are generated by a neural-network prior rather than being sampled uniformly. Using statistical physics techniques, we derived a Bayesian inference algorithm which we conjecture to be asymptotically optimal among polynomial-time algorithms, when considering locally tree-like graphs and i.i.d. Gaussian features. 
We have analysed the performance of the resulting BP-AMP algorithm against estimators that use only information from the spreading (BP-only) or that use only the covariate variables (AMP-only), finding a great advantage overall in combining the two techniques.
For binary weights, our analysis revealed a statistical-to-computational gap, driven by first-order phase transitions to perfect recovery, as a function of the problem's signal-to-noise ratio. This phenomenology, absent in models with uniform priors, highlights how a neural network model for the sources' prior can drastically alter the inference problem's hardness.

Several future directions stem from this work. We assumed all model parameters were known; a natural extension is to study the more realistic case where these parameters must be learned.

Future work could also explore more complex neural-network priors, such as multi-layer architectures~\cite{aubin2020exact}. This would technically require combining results from~\cite{manoel2017multi} and~\cite{ghio2023bayes}, to study the influence of network depth. Alternatively, the neural prior could be applied directly to the transition times, which would necessitate a multi-class output AMP algorithm as derived in~\cite{Cornacchia2023learning}.

Finally, developing rigorous mathematical proofs for the more general hardness conjectures presented here remains an important open question.

\section*{Acknowledgments}
We thank Odilon Duranthon for useful discussions.
This research was supported by the Swiss National Science Foundation (SNSF) under grant numbers 212049 (SMArtNet) and 200390 (OperaGOST).

\newpage

\appendix

\section{Derivation of Algorithm~\ref{Algo_GLM}\label{App:GLM_deriv}}
We first review the cavity equations for the inference on the spreading process and for the the neural-network prior separately, before showing how they are combined.

\subsection{Belief propagation for the spreading process}\label{App:BP}
This section reports the belief propagation (BP) equations for spreading models with the probability function in~\eqref{eq:Spread1}. The BP equations for these models were first derived in~\cite{altarelli2014bayesian}. We adopt the notation of~\cite{ghio2023bayes}, which is more directly adaptable to the present case.

The variables $\vect{T}_{(i,j)}$ in Fig.~\ref{fig:FactorG_GLM} compactly represent the pair of variables $\{\vect{t_i},\vect{t_j}\}$ on a modified factor graph that preserves the local tree-like structure. The detailed derivation can be found in~\cite{ghio2023bayes}; here we give the final form of the message-passing equations:
\begin{equation}\label{eq:BP_gen}
    m_{i\rightarrow j}(\vect{t_i
    },\vect{t_j}) =
    \frac{1}{Z_{i\rightarrow j}}\sum_{x_i^0}
    \sum_{ \{\vect{t_k} \}_{k\in \partial_i\backslash j}}
   \widetilde{\Psi}_i(\vect{t_i}, \{\vect{t_j}\}_{j\in\partial_i},x_i^0,\mathcal{O}_i, \vect{\Theta})
 \prod_{k\in\partial_i\backslash j}
  m_{k\rightarrow i} ( \vect{t_k},\vect{t_i})\,,
\end{equation}
from which one can compute the marginal for each variable as
\begin{align}\label{eq:BPmarg}
    b_i\left(\vect{t_i}\right) &= \frac{1}{Z_i}
    \sum_{x_i^0}\sum_{ \{\vect{t_k} \}_{k\in \partial_i}}
    \widetilde{\Psi}_i(\vect{t_i}, \{\vect{t_j}\}_{j\in\partial_i},x_i^0,\mathcal{O}_i, \vect{\Theta})
 \prod_{k\in\partial_i}
  m_{k\rightarrow i} ( \vect{t_k},\vect{t_i}) = \\
  &= \frac{Z_{i\rightarrow j}}{Z_i}\sum_{\vect{t_j}} m_{j\rightarrow i} ( \vect{t_j}, \vect{t_i})m_{i\rightarrow j} (\vect{t_i}, \vect{t_j})\,,
\end{align}
where $Z_i$ is the normalization. The free entropy for a single instance can then be written as
\begin{equation}
    \frac{1}{N}\log Z = \frac{1}{N}\sum_{i=1}^N \log Z_i - \frac{1}{N} \sum_{(i,j) \in E}\log Z_{(i,j)}\,,
\end{equation}
where $Z_{(i,j)} = \sum_{\vect{t_i},\vect{t_j}} m_{j\rightarrow i} ( \vect{t_j},\vect{t_i})m_{i\rightarrow j} ( \vect{t_i},\vect{t_j})$.

\subsection{Approximate message passing for the Perceptron prior}
Approximate message passing (AMP) is a message-passing algorithm tailored for inference on dense graphical models, such as the Generalized Linear Models (GLMs)~\cite{zdeborova2016statistical} that include the one-layer perceptron prior used here. We sketch its derivation, adapting the notation to our setting.

The GLM prior generates the initial states $\vect{x^0}$ from the vector of weights $\vect{u}\!\in\!\mathbb{R}^{M}$ (with prior $P^{u}$) and a random matrix $F \in \mathbb{R}^{N\times M}$ (with $F_{ia}\!\sim\! \mathcal N(0,1/M)$) via a perceptron channel:
\begin{equation}
    P_{\rm out}(x_i^0 |z_i ) = \delta \Bigl( x_i^0 -\mathrm{sign}(z_i -\kappa) \Bigr) \qquad \text{where} \qquad z_i \;=\; \sum_{a=1}^{M}\!F_{ia}\,u_{a}\,.
\end{equation}
The factor graph for this prior is dense. In the thermodynamic limit ($M,N\to \infty$ with $\alpha = N/M = O(1)$), the cavity fields acting on any single variable are sums of many weakly correlated terms. By the central-limit theorem, these fields concentrate and their distributions become Gaussian, allowing them to be characterised by just their first two moments.

The key messages in the dense sub-graph of Fig.~\ref{fig:FactorG_GLM} are $\chi_{i\to a}(u_a)$ (from node $i$ to the weight $a$) and $\bar\chi_{a\to i}(u_a)$ (from weight $a$ to node $i$). Due to the concentration argument, these messages can be approximated as Gaussian distributions. Specifically, the message $\bar\chi_{a\to i}(u_a)$ takes the form:
\begin{equation}
  \bar\chi_{a\to i}(u_a)
  \;\propto\;
  P^u(u_a)\,
  \exp\!\Bigl[\!-\,\tfrac{A}{2}\,u_a^2 + B_{a\to i}\,u_a\Bigr],
\end{equation}
where $A$ and $B_{a\to i}$ are effective parameters representing the variance and mean of the cavity field on $u_a$. This Gaussian approximation is the foundation of the AMP algorithm.

The algorithm proceeds by iteratively updating estimates for the means ($a_a$) and variances ($v_a$) of the weights, and the effective fields ($\omega_i$) acting on the output nodes. This is achieved through scalar denoising functions for the input ($f_a, f_v$) and output ($g_o$) channels. The standard AMP updates for a GLM take the form:
\begin{align}\label{eq:AMP-GLM}
  V^{(t+1)}      &= \frac1M\sum_{a} v_a^{(t)}, \\
  \omega_i^{(t+1)} &= \sum_{a} F_{ia}a_a^{(t)} -V^{(t+1)} g_{o,i}^{(t)}, \\
  g_{o,i}^{(t+1)} &= g_o\bigl(\omega_i^{(t+1)},x_i^0,V^{(t+1)}\bigr), \\
  A^{(t+1)}       &= \frac1M\sum_{i} \bigl(g_{o,i}^{(t+1)}\bigr)^2 , \\
  B_a^{(t+1)}&= \sum_{i} F_{ia} g_{o,i}^{(t+1)} + a_a^{(t)} A^{(t+1)}, \\
  a_a^{(t+1)}&= f_a\!\bigl(A^{(t+1)},B_a^{(t+1)}\bigr),\;
  v_a^{(t+1)} = f_v\!\bigl(A^{(t+1)},B_a^{(t+1)}\bigr).
\end{align}
Here, the terms $g_{o,i}^{(t)}$ in the update for $\omega_i$ and $a_a^{(t)} A^{(t+1)}$ in the update for $B_a$ are the crucial Onsager reaction terms. They correct for the ``self-interaction'' that would otherwise arise from using a variable's estimate to compute the field acting back on it, ensuring the algorithm's dynamics can be tracked accurately by State Evolution~\cite{barbier2019optimal}. The functions $g_o$, $f_a$, and $f_v$ are the scalar denoisers defined in the main text.

\subsection{Putting the two together}
In the NSS model, the posterior factorizes into two sub-problems: a dense GLM part coupling $(\vect u,\vect x^0)$ and a sparse spreading part coupling $(\vect x^0,\{\vect t_i\})$. This structure motivates a hybrid message-passing architecture: AMP handles the dense GLM part, providing prior messages $\eta_i(x_i^0)$ to BP, which in turn computes refined beliefs $\nu_i(x_i^0)$ that act as effective likelihoods for AMP.

Concretely, at each iteration $t$ we perform two steps:
\begin{itemize}
\item \textbf{BP-step}: Update all epidemic messages $m_{i\to j}^{(t)}(\vect t_i,\vect t_j)$ using~\eqref{eq:BP_gen}, where the source term is now weighted by the message $\eta_i^{(t)}(x_i^{0})$ from AMP. From the updated messages, compute the new marginals on the initial states:
      \begin{equation}
          \nu^{(t+1)}_i(x_i^0) = \sum_{t_i,\vect{t_{k\in\partial i}}}\widetilde{\psi}_i(t_i,\vect{t_{k\in\partial i}} ,\mathcal{O}_i, x_i^0)\prod_{k\in\partial i}m^{(t)}_{k\rightarrow i}(t_k,t_i)\,.
      \end{equation}
\item \textbf{AMP-step}: Use the BP marginals $\nu_i^{(t+1)}(x_i^{0})$ as effective, data-dependent likelihoods in the GLM–AMP update. The resulting field, to be passed back to BP, is computed as:
      \begin{equation}\label{eq:eta_GLM_final}
        \eta^{(t+1)}_i(x_i ^0 ) \propto \int dz P_{\rm out}(x_i ^0 |z) \exp\Bigl[ -\frac{(z-\omega_i^{(t+1)} )^2}{2V^{(t+1)}} \Bigr]\,.
      \end{equation}
\end{itemize}
To merge the two algorithms, we only need to modify the output denoising function $g_o$ to incorporate the prior knowledge $\nu_i(x_i^0)$ from the BP step:
\begin{equation}
  g_o(\omega,\nu,V)\;=\;
  \frac{\displaystyle \int\!dz \sum_{x^0}\nu(x^0)\,P_{\rm out}(x^0|z)\,(z-\omega)\;e^{-\frac{(z-\omega)^2}{2V}}}{\displaystyle V\int\!dz \sum_{x^0}\nu(x^0)\,P_{\rm out}(x^0|z)\; e^{-\frac{(z-\omega)^2}{2V}}}\,.
\end{equation}
With this modification, the AMP updates in~\eqref{eq:AMP-GLM} and the BP updates in~\eqref{eq:BP_gen} (now weighted by $\eta_i$) are iterated, forming the complete Algorithm~\ref{Algo_GLM}. This construction is inspired by~\cite{duranthon2023neural,ghio2023bayes}. The limit $F\to0$ recovers the BP-only inference from~\cite{ghio2023bayes}, while the limit $\lambda_{ij}\to0$ collapses to the standard perceptron-AMP from~\cite{zdeborova2016statistical}. In practice, damping and asynchronous updates are used to ensure convergence.

\section{\label{app:FreeEnt}Free-Entropy}
Given a probability distribution, and the associated factor graph, the Bethe approximation of the log–partition function is given by
\begin{equation}
\label{eq:bethe_general}
\log Z_{\rm Bethe} = \sum_{a} \phi_a - \sum_{(a,b)} \phi_{ab}\,,
\end{equation}
where the sum over $a$ runs over all factor nodes and variable nodes, and $(a,b)$ runs over all edges in the factor graph. We can decompose the free entropy into three pieces:
\begin{equation}\label{eq:PhiRS_app}
\phi_{\rm RS} \equiv \frac{1}{N}\log Z_{\rm Bethe} = \phi_{\rm spread} + \phi_{\rm GLM} + \phi_{\rm inter}\,,
\end{equation}
with
\begin{align}
\phi_{\rm spread} &= \frac{1}{N} \sum_{i=1}^N \phi_i^{\rm spread} + \frac{1}{N} \sum_{(i,j)\in E} \phi_{(i,j)}^{\rm spread} - \frac{1}{N} \sum_{(i,j)\in E} \phi_{i,(i,j)}^{\rm spread}- \frac{1}{N} \sum_{(i,j)\in E} \phi_{j,(i,j)}^{\rm spread}\,,\label{eq:phi_spread}\\[1mm]
\phi_{\rm GLM} &= \frac{1}{N} \sum_{i=1}^N \phi_i^{\rm out} + \frac{1}{N} \sum_{a=1}^M \phi_a - \frac{1}{N} \sum_{i=1}^N\sum_{a=1}^M\phi_{ia}\,,\label{eq:phi_glm}\\[1mm]
\phi_{\rm inter} &= \frac{1}{N} \sum_{i=1}^N \phi_i^{0} - \frac{1}{N} \sum_{i=1}^N \phi_i^{0, {\rm out}} - \frac{1}{N} \sum_{i=1}^N \phi_i^{0, {\rm spread}}\,. \label{eq:phi_inter}
\end{align}

As done in~\cite{ghio2023bayes}, one can show that $\phi_{i,(i,j)}^{\rm spread} = \phi_{j,(i,j)}^{\rm spread} = \phi_{(i,j)}^{\rm spread}$, so that for the spreading part we have
\begin{equation}\label{eq:PhiSpread}
    \phi_{\rm spread} =  \frac{1}{N} \sum_{i=1}^N \phi_i^{\rm spread} - \frac{1}{N} \sum_{(i,j)\in E} \phi_{(i,j)}^{\rm spread}\,,
\end{equation}
where 

\begin{equation}
    \phi_i^{\rm spread} =  \log \sum_{x_i^0,t_i,\{t_j\}_{j\in\partial i}} \eta_i(x_i^0)\widetilde{\psi}_i\Bigl(x_i^0,t_i,\{t_j\}_{j\in\partial i},\mathcal{O}_i, \bm{\lambda},F_i,\mathbf{u}\Bigr)
    \prod_{j\in\partial i} m_{j\to i}(t_j,t_i)\,,
\end{equation}
and
\begin{equation}
    \phi_{(i,j)}^{\rm spread} = \log \sum_{t_i,t_j} m_{i\to j}(t_i,t_j)\, m_{j\to i}(t_j,t_i)\,.
\end{equation}

By the same argument, one can show that $\phi_i^{0} = \phi_i^{0, {\rm out}} = \phi_i^{0, {\rm spread}}$ and thus
\begin{equation}
    \phi_{\rm inter} = - \frac{1}{N} \sum_{i=1}^N \phi_i^{0}\,,
\end{equation}
in which
\begin{align}
     \phi_i^{0} &= \log \sum_{x_i^0} \eta_i(x_i^0)\,\nu_i(x_i^0) \\
     &= \log \left(\frac{1}{Z_i^{\nu}}\sum_{x_i^0} \eta_i(x_i^0)\sum_{t_i,\vect{t_{k\in\partial i}}}\widetilde{\Psi}_i(t_i,\vect{t_{k\in\partial i}} , x_i^0,\mathcal{O}_i)\prod_{k\in\partial i}m_{k\rightarrow i}(t_k,t_i)\right) \\
     &= \phi_i^{\rm spread} -\log Z_i^{\nu}\,,
     \label{eq:phi0}
\end{align}
where we made the message $\nu_i$ explicit. It is easy to see that in the total free entropy in~\eqref{eq:PhiRS_app}, the first term in~\eqref{eq:phi0} cancels the first term in~\eqref{eq:PhiSpread}.

Finally, for the GLM part we have
\begin{align}
    \phi_a &= \log \left(\int \mathrm{d}u_a\,\psi_{a}(u_{a})\prod_{j}\chi_{j\rightarrow a}(u_{a})\right) \\
    &= \log \left(\int \mathrm{d}u_a\,\psi_{a}(u_{a})\prod_{j}\left(\frac{1}{Z_{j\rightarrow a}}\sum_{x_j^0}\nu_j(x_j^0)\int\left(\prod_{b\neq a}{\rm d}u_{b}\right)\Psi_j^{\rm out}(x_j^0,\vect{u},\vect{F_j})\prod_{b\neq a}\bar{\chi}_{b\rightarrow j}(u_{b})\right)\right) \\
    &= -\sum_j \log Z_{j\rightarrow a} + \log Z_{a}\,,
\end{align}
and
\begin{equation}
    \phi_i^{\rm out} = \log\left(\sum_{x_i^0}\nu_i(x_i^0)\int\left(\prod_{b}{\rm d}u_{b}\right)\Psi_i^{\rm out}(x_i^0,\vect{u},\vect{F_i})\prod_{b}\bar{\chi}_{b\rightarrow i}(u_{b})\right)\,,
\end{equation}
while
\begin{align}
    \phi_{ia} &= \log\left(\int\dd u_{a} \bar{\chi}_{a \rightarrow i}(u_{a})\chi_{i \rightarrow a}(u_{a})\right) \\
    &= \log\left(\frac{1}{Z_{i\rightarrow a}}\sum_{x_i^0}\nu_i(x_i^0)\int\left(\prod_{b}{\rm d}u_{b}\right)\Psi_i^{\rm out}(x_i^0,\vect{u},\vect{F_i})\prod_{b}\bar{\chi}_{b\rightarrow i}(u_{b})\right) \\
    &= \phi_i^{\rm out} - \log Z_{i\rightarrow a}\,.
\end{align}

Putting all together, we get
\begin{equation}
    \phi_{\rm RS} = \frac{1}{N}\sum_{i=1}^N \log Z_i^{\nu}  - \frac{1}{N} \sum_{(i,j)\in E} \phi_{(i,j)}^{\rm spread} + \frac{1}{N}\sum_{a=1}^M \log Z_{a} + \frac{1-M}{N}\sum_{i=1}^N\phi_i^{\rm out}\,.
\end{equation}
One can show, see for example~\citep[Appendix B]{duranthon2023neural}, that 
\begin{align}
    \sum_{a=1}^M \log Z_{a} = M&\sum_{i=1}^N \phi_i^{\rm out} + \sum_{a=1}^M \log \int \mathrm{d}u_a\,P^u(u_a)\,\exp\Bigl[-\frac{A_a}{2}u_a^2 + B_a u_a\Bigr]\\
    &+\sum_{a=1}^M \Bigl[\frac{A_a}{2}(a_a^2+v_a) - B_a\,a_a\Bigr] + \sum_{i=1}^N \frac{\Bigl(\omega_i - \sum_{a=1}^M F_{ia}a_a \Bigr)^2}{2V_i}\,.
\end{align}
To conclude, the final formula for the RS free entropy reads
\begin{align}
    \phi_{\rm RS} = \frac{1}{N}&\sum_{i=1}^N \log Z_i^{\nu}  - \frac{1}{N} \sum_{(i,j)\in E} \phi_{(i,j)}^{\rm spread}  + \sum_{a=1}^M \log \int \mathrm{d}u_a\,P^u(u_a)\,\exp\Bigl[-\frac{A_a}{2}u_a^2 + B_a u_a\Bigr]\\
    &+\frac{1}{N}\sum_{i=1}^N \phi_i^{\rm out}+\sum_{a=1}^M \Bigl[\frac{A_a}{2}(a_a^2+v_a) - B_a\,a_a\Bigr] + \sum_{i=1}^N \frac{\Bigl(\omega_i - \sum_{a=1}^M F_{ia}a_a \Bigr)^2}{2V_i}\,,
\end{align}
and it can be computed in linear time by using the BP messages and AMP variables in Algorithm~\ref{Algo_GLM}.
\begin{figure}
    \centering
    \includegraphics[width=0.85\linewidth]{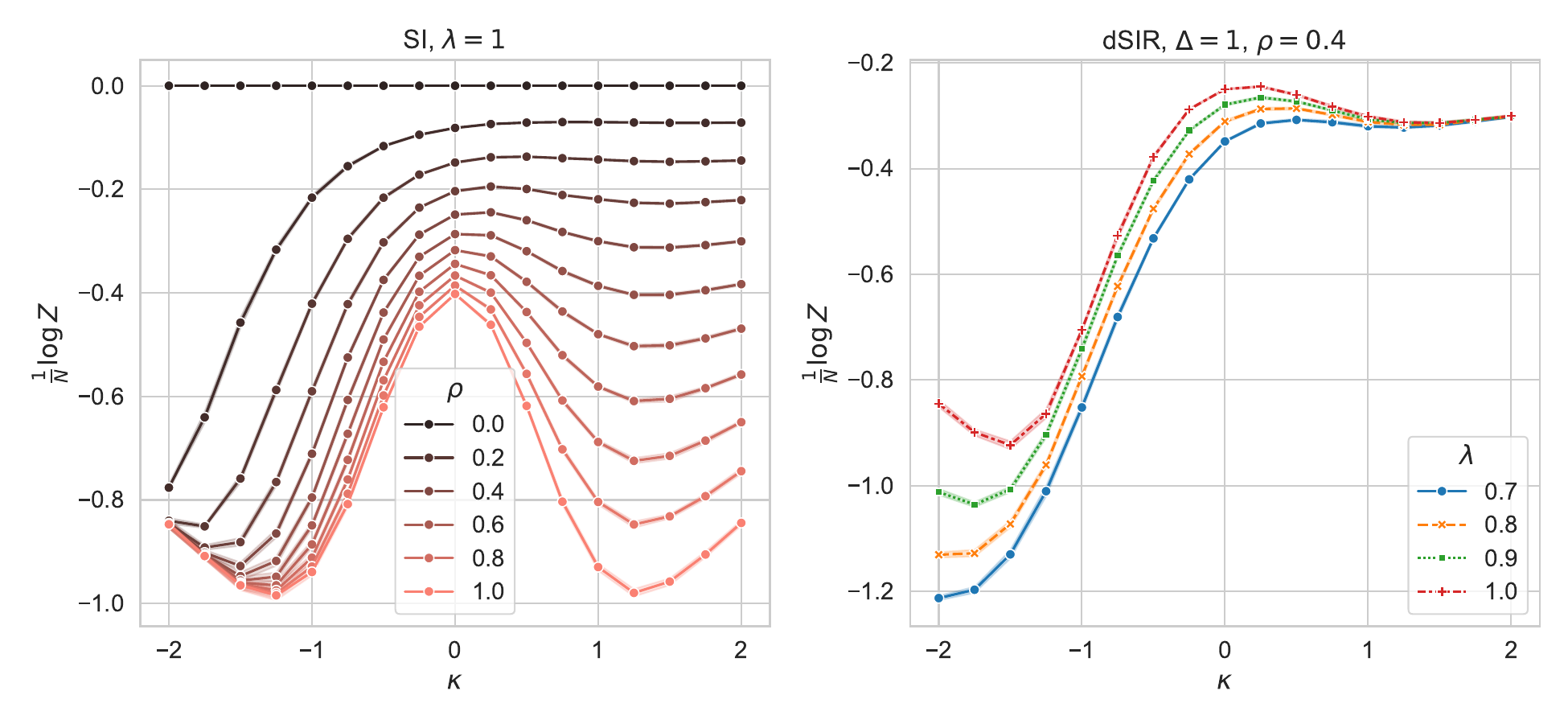}
    \caption{\textbf{Free entropy with Gaussian weights.} PSS model, Gaussian weights, $\alpha=4$. The x-axis varies the threshold $\kappa$, changing the source fraction $\delta$. We plot the total free entropy for sensor observations. Left: SI model ($\lambda=1$) with varying sensor fraction $\rho$. Right: dSIR model ($\Delta=1$) with fixed $\rho=0.4$ and varying transmission probability $\lambda$. Simulations on RRGs ($d=3, N=20000$). Each point is an average of 20 runs; shading is the 99\% confidence interval. Results from random and informed initialisations are indistinguishable.}\label{fig:logZgauss}
\end{figure}

Fig.~\ref{fig:logZgauss} shows the free entropy for the Gaussian weights case, where no first-order phase transitions are observed. The left panel plots the free entropy as a function of the perceptron threshold $\kappa$, which controls the source density. While the profile is symmetric around $\kappa=0$ for $\rho=0$ (no observations) and $\rho=1$ (full observation), it becomes asymmetric for intermediate sensor fractions. The right panel shows that spreading stochasticity (varying $\lambda$) has a relatively minor impact on the free entropy, even for the dSIR model with $\Delta=1$, where its effect should be most pronounced.

Finally, we discuss the calculation of the free entropy at the informative fixed point, $\phi_\mathrm{info}$. For the non-stochastic SI model ($\lambda=1$), recovering the sources is equivalent to recovering the full infection trajectories. This corresponds to a state where all BP messages are delta functions centered on the ground-truth values. In this simplified scenario, the only non-trivial contribution to the free entropy comes from the prior over the weights, yielding:
\begin{equation}\label{eq:Phi_info}
    \phi_\mathrm{info} (\lambda=1) = \frac{1}{\alpha}\mathbb{E}_u \log P^u(u).
\end{equation}
For the stochastic case ($\lambda\neq1$), the situation is more complex, as perfect source recovery no longer implies perfect trajectory recovery. To compute $\phi_\mathrm{info}$ in this setting, we run the algorithm with an additional observation of the true initial state for all nodes. The algorithm then infers the subsequent spreading given the known sources, and the free entropy at the resulting fixed point gives the value of $\phi_\mathrm{info}$.

\section{\label{app:AddPlot}Additional plots}
\subsection{\label{sec:Nishi}Nishimori conditions for Gaussian weights}
\begin{figure}
    \centering
    \includegraphics[width=0.85\linewidth]{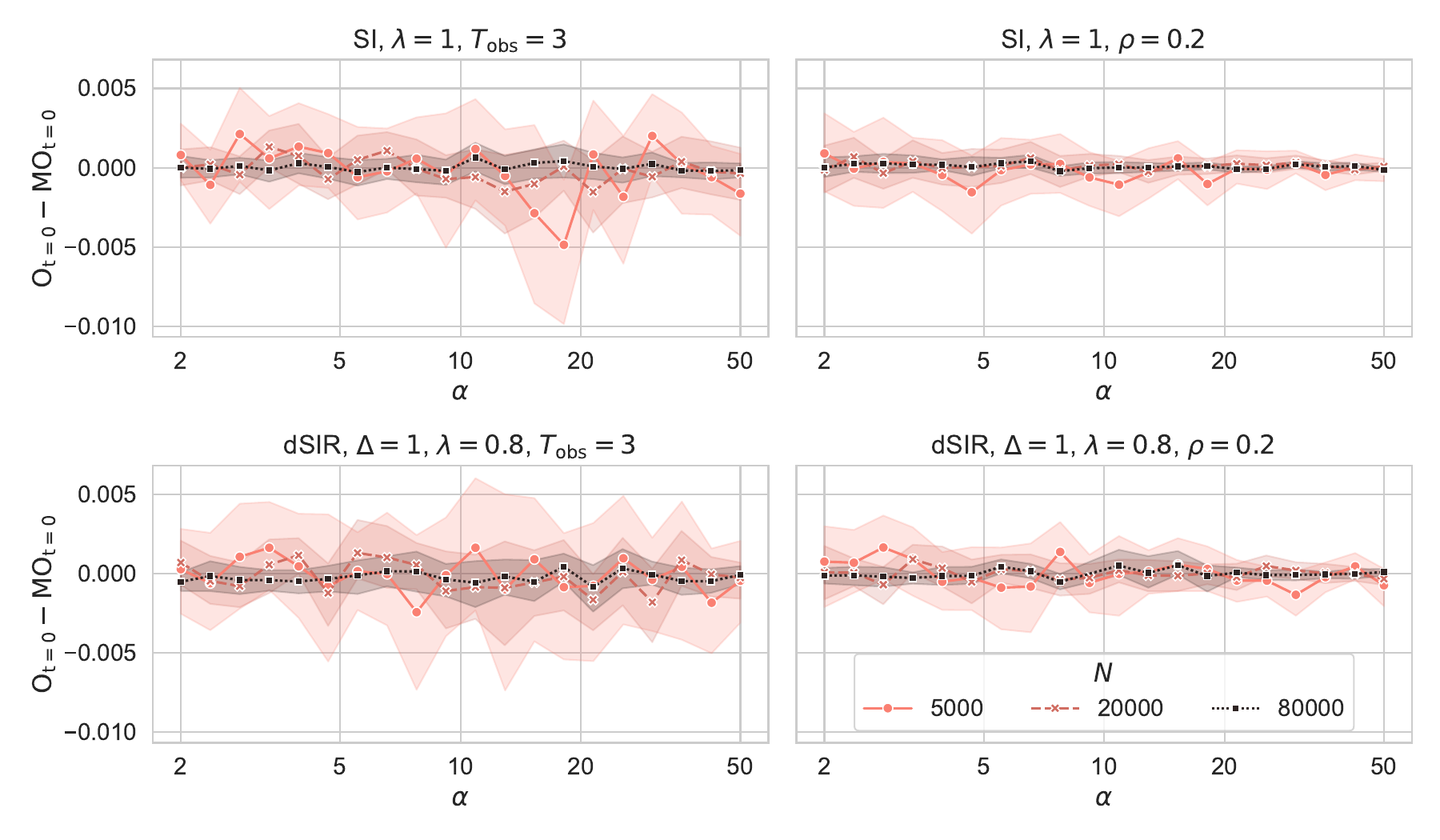}
    \caption{\textbf{Check of the Nishimori conditions.} We consider the PSS model with $\kappa=-1$ and Gaussian weights. We plot the difference between overlap and mean overlap at $t=0$ as a function of $\alpha\in[2,50]$, in logarithmic scale. In the left panels, we consider snapshot observations at time $T_{\rm obs} = 3$, while on the right we look at the case in which there is a fraction $\rho=0.2$ of sensors. At the same time, in the upper panel we look at SI model with transmission parameter $\lambda=1$, while in the lower panels we consider the dSIR model with $\Delta=1$ and $\lambda=0.8$. All simulations are done on RRGs of degree $d=3$, and we vary the system size $N\in[5000,80000]$ to check for finite size effects. Each point in the plot has been computed by averaging 20 different simulations, and the shaded region represents the $99\%$ confidence interval.}\label{fig:Nishi}
\end{figure}
In this section, we check that the Nishimori conditions are satisfied in the case of Gaussian weights. In Fig.~\ref{fig:Nishi}, we fix $\kappa = -1$, so that on average the fraction of sources is $\delta = \frac{1}{2}(1-\text{erf}(1/\sqrt{2})) \approx 0.159$. We then explore the parameter space of our problem, and we find that in all cases, on average, the Nishimori conditions are satisfied. We plot the curves for different values of the system size $N$ and, as expected from self-averaging properties, we see that the difference between the overlap and the mean overlap diminishes when we increase $N$.

\subsection{\label{app:PhT} Phenomenology of the first order transition for Rademacher variables}
\begin{figure}
    \centering
    \includegraphics[width=0.8\linewidth]{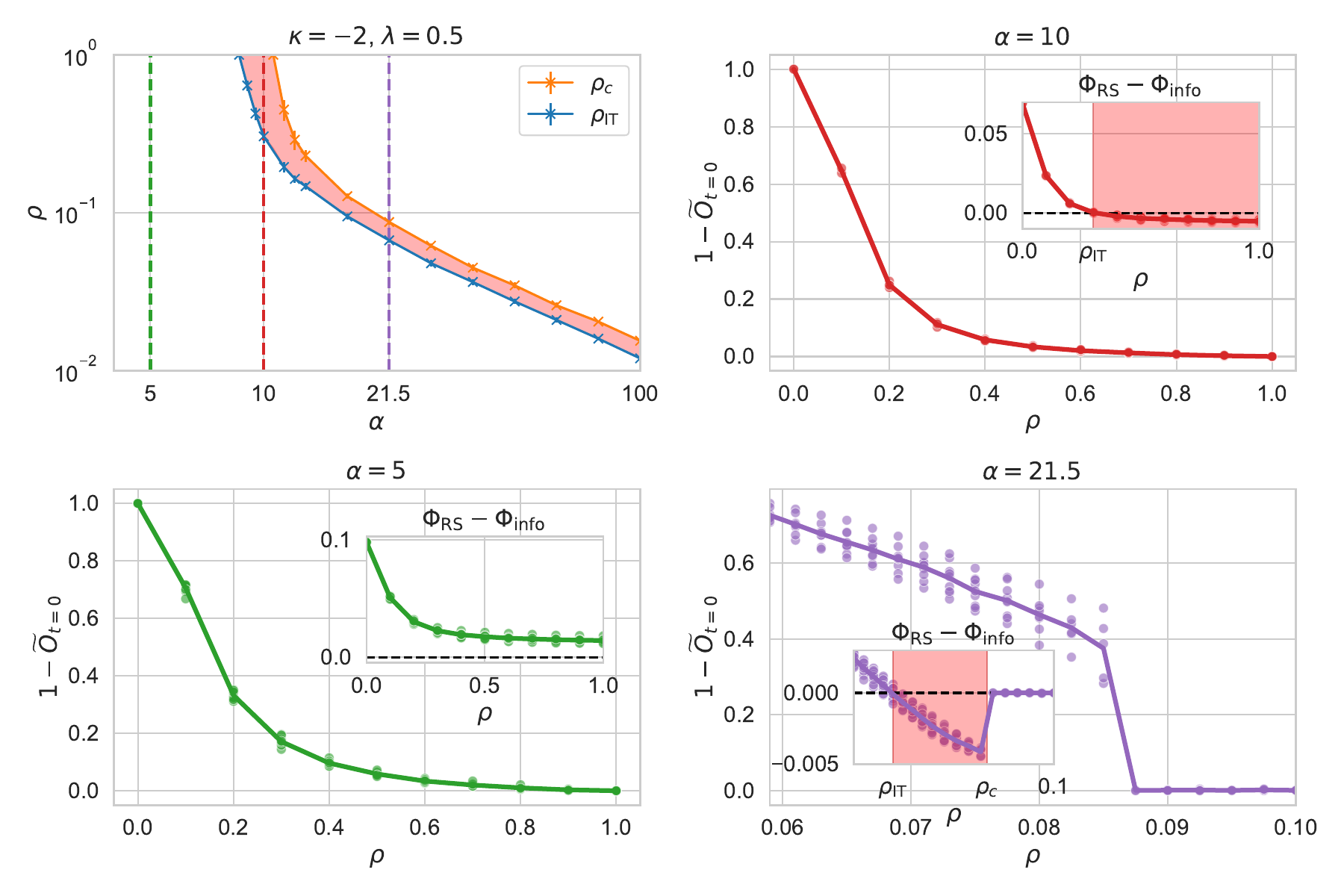}
    \caption{\textbf{Phase transition phenomenology.} We fix $\kappa=-2$ (corresponding to $\delta \approx0.023$), and $\lambda=0.5$. Top left: $(\rho,\alpha)$ phase diagram. Other plots show free energy difference and $1-\widetilde{O}_{\rm t=0}$ vs. $\rho$ for fixed $\alpha$ in each of the three regimes. We see that, from low to high $\alpha$, we first have no phase transition, then we encounter a hard phase but do not have perfect recovery, and finally we have both the hard phase and the perfect recovery phase.}\label{fig:phase_diag_sum}
\end{figure}

In this section, we present an example of what happens when changing the ratio $\alpha$ in the phase diagram for Rademacher variables, presented at the end of Section~\ref{sec:Nishi}. Specifically, in Fig.~\ref{fig:phase_diag_sum} we take the case $\kappa=-2$ and $\lambda=0.5$, and we show the behaviour of the overlap and Free energy as a function of $\rho$ for three cases: 1) $\alpha=5<\alpha_{\rm IT}^{\rm AMP}$, in which there is no hard phase and no transition, and thus the algorithm is always Bayes-optimal. 2) $\alpha_{\rm IT}^{\rm AMP}<\alpha=10<\alpha_{\rm c}^{\rm AMP}$, where again there is no transition to perfect recover but now for $\rho>\rho_{\rm IT}$ we see a hard phase by looking at the free energy. 3) $\alpha > \alpha_{\rm c}^{\rm AMP}$, in which we finally have a transition to perfect recovery for $\rho= \rho_{\rm c}$.
\subsection{\label{sec:MSE}Mean squared error}
In the main text, we focused on the problem of retrieving the sources of the spreading, for convenience. However, using a Bayesian approach allows us to characterise the performance of our algorithm in very generic tasks, just by using the marginals estimated through the belief propagation and approximate message passing algorithms.
For example, it is interesting to look at how well the algorithm is able to recover the entire trajectory of each individual, which for $\lambda<1$, where the spreading is stochastic, is a problem strictly harder than source recovering. If we restrict to models in which there is a single transition time, like the SI and dSIR models we described in the main text, we can use as performance parameter the Squared Error between the ground truth transmission times $\vect{t^*} \equiv\{t_i^*\}_{i=1}^N$ and the ones estimated through the algorithm $\vect{\hat{t}} \equiv \{\hat{t}_i\}_{i=1}^N$, defined as
\begin{equation}
    \rm{SE}(\vect{t^*},\vect{\hat{t}}) \equiv \frac{1}{N}\sum_{i=1}^N(\hat{t}_i-t_i^*)^2\,.
\end{equation}
As for the overlap estimator, we assume that the algorithm has additional information on the process through partial observations and feature covariates, so that the Bayes-optimal estimator uses the posterior probability distribution $P(\vect{t}\mid\vect{\mathcal{O}},F)$ and evaluates the Mean Squared Error, defined as
\begin{equation}
    \rm{MSE}(\vect{\hat{t}}) \equiv \mathbb{E}_{\vect{t}\mid\vect{\mathcal{O}},F}\left[\rm{SE}(\vect{t},\vect{\hat{t}})\right] = \frac{1}{N}\sum_{\vect{t}}P(\vect{t}\mid\vect{\mathcal{O}},F)\sum_{i=1}^N (\hat{t}_i-t_i)^2\,.
\end{equation}
In this case, in order to maximise this quantity over $\vect{\hat{t}}$, one can show~\cite{zdeborova2016statistical} that we have to consider the minimum mean-squared error estimator:
\begin{equation}\label{eq:MMSE}
    \hat{t}_i^{\rm{MMSE}} = \sum_{\vect{t}}t_iP(\vect{t}\mid \vect{\mathcal{O}},F)= \sum_{t_i}t_iP_i(t_i\mid \vect{\mathcal{O}},F)\quad \forall\,i\in[1:N]
\end{equation}
where we have defined the marginal probability $P_i(t_i\mid \vect{\mathcal{O}},F) \equiv \sum_{\{t_j\}_{j\neq i}}P(\vect{t}\mid\vect{\mathcal{O}},F)$. In this section, we will use as performance parameters the rescaled squared error
\begin{equation}\label{eq:res_se}
    \rm{R}_{\rm SE} = \frac{\rm{SE}(\vect{t^*},\vect{\hat{t}}^{\rm RND}) - \rm{SE}(\vect{t^*},\vect{\hat{t}}^{\rm MMSE})}{\rm{SE}(\vect{t^*},\vect{\hat{t}}^{\rm RND}) }\,,
\end{equation}
and the associated rescaled mean squared error
\begin{equation}
    \rm{R}_{\rm MSE} = \frac{\rm{MSE}(\vect{\hat{t}}^{\rm RND}) - \rm{MSE}(\vect{\hat{t}}^{\rm MMSE})}{\rm{MSE}(\vect{\hat{t}}^{\rm RND}) }\,,
\end{equation}
where $\hat{t}_i^{\rm RND} = \sum_{t_i}P_i\giventhat{t_i}{\emptyset,F}t_i$ is the random estimator. 

Notice that these parameters are equal to zero if the MMSE estimator and the RND estimator have the same performance, and they are equal to one if the MMSE estimator recovers all the trajectories perfectly. As done for the overlap, we can probe the optimality of the algorithm by checking the Nishimori conditions, i.e.\ the difference between $\rm{R_{SE}}$ and $\rm{R_{MSE}}$, since from the Nishimori conditions we know that they coincide on average in the Bayes-optimal setting.
\begin{figure}
    \centering
    \includegraphics[width=1\linewidth]{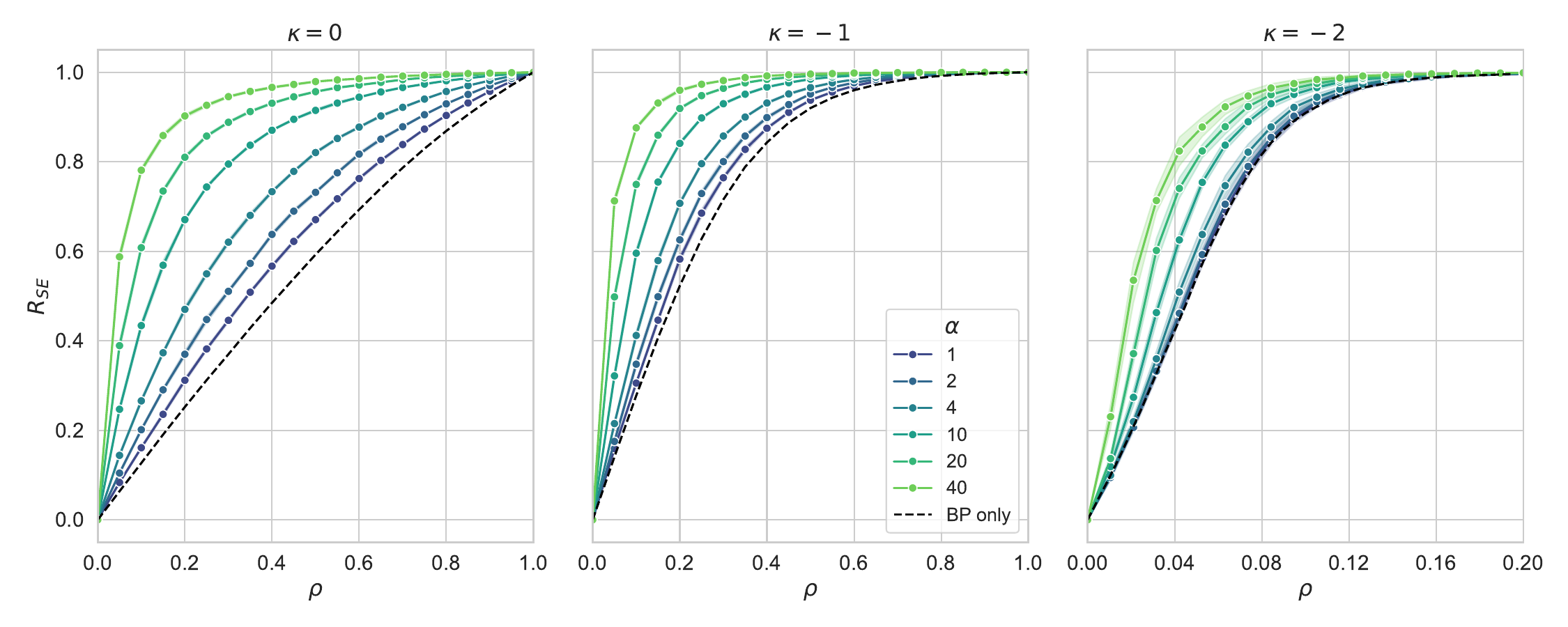}
    \caption{\textbf{Squared Error (Gaussian).} We look at the PSS model with Gaussian weights, varying the ratio $\alpha=N/M$. We also choose the threshold value $\kappa=0,-1,-2$, corresponding to an average fraction of sources $\delta \approx0.5, 0.159, 0.023$ respectively. We consider observations through sensors, considering the SI model with $\lambda=1$. All simulations are done on RRGs of degree $d=3$ and size $N=20000$. We plot the rescaled squared error defined in~\eqref{eq:res_se} as a function of the fraction of sensors $\rho\in[0,1]$. Each point in the plot has been computed by averaging 20 different simulations, and the shaded region represents the $99\%$ confidence interval.}\label{fig:PhD_Gauss_SE}
\end{figure}
\begin{figure}
    \centering
    \includegraphics[width=1\linewidth]{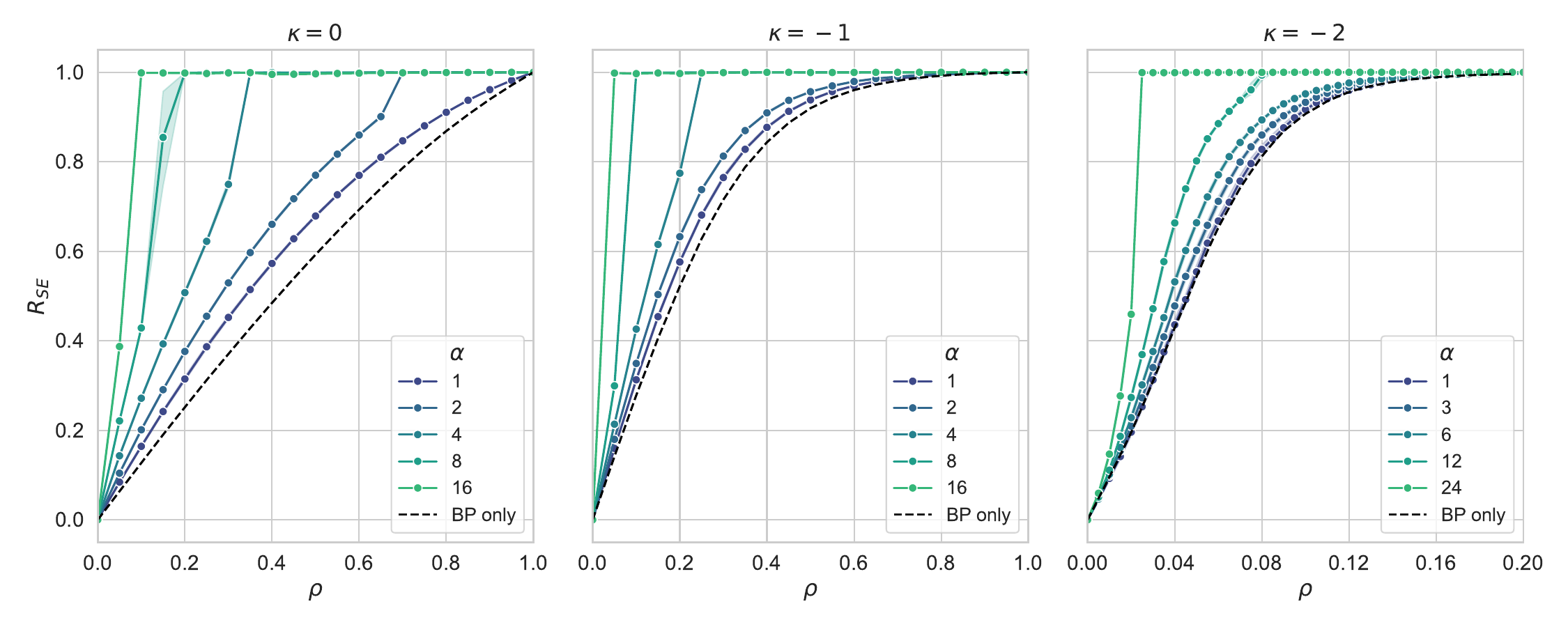}
    \caption{\textbf{Squared Error (Rademacher).} We look at the PSS model with Rademacher weights, varying the ratio $\alpha=N/M$. We also choose $\kappa=0,-1,-2$, corresponding to an average fraction of sources $\delta \approx 0.5, 0.159, 0.023$ respectively. We consider observations through sensors, considering the SI model with $\lambda=1$. All simulations are done on RRGs of degree $d=3$ and fixing the size $N$ in such a way that $M* N = 1.6\times10^9$. We plot the rescaled squared error defined in~\eqref{eq:res_se} as a function of the fraction of sensors $\rho\in[0,1]$. Each point in the plot has been computed by averaging 20 different simulations, and the shaded region represents the $99\%$ confidence interval.}\label{fig:PhD_Rade_SE}
\end{figure}
In Fig.~\ref{fig:PhD_Gauss_SE} and~\ref{fig:PhD_Rade_SE} we show the performance of the algorithm in terms of the squared error, in the same settings as Fig.~\ref{fig:PhD_Gauss} and~\ref{fig:PhD_Rade} in the main text. We see again that for $\alpha$ going to zero we retrieve the performance of BP, while the more we increase the correlations between the weights (by increasing $\alpha$) the more the performance develops a gap from the BP-only baseline.

\subsection{Finite size study for Rademacher variables}
In this section, we discuss the behaviour of the Nishimori conditions when considering Rademacher variables and values of $\rho$ near the transition to perfect recovery.

\begin{figure}
    \centering
    \includegraphics[width=0.7\linewidth]{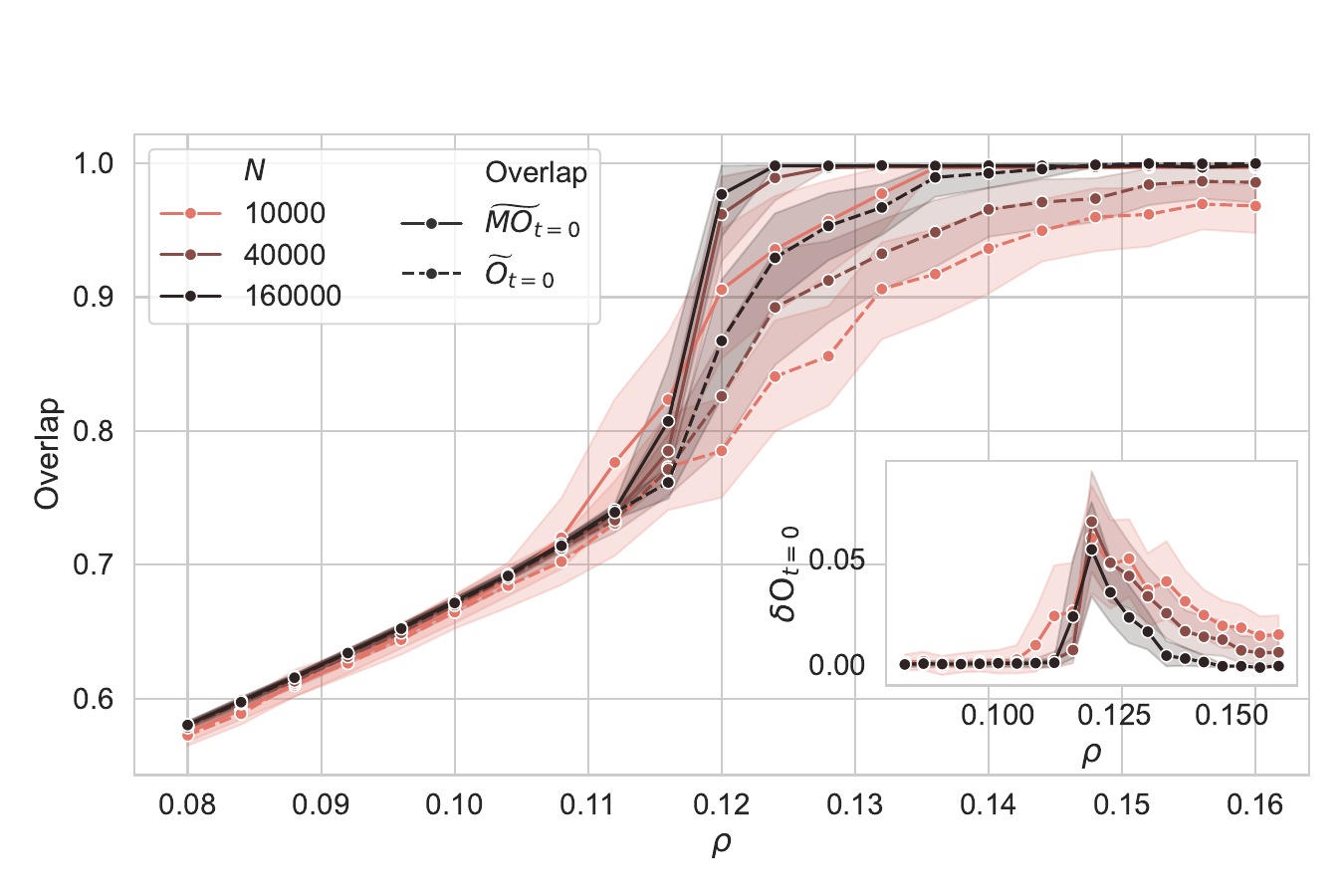}
    \caption{\textbf{Nishimori conditions and finite sizes.} We consider the PSS model with Rademacher weights, fixing $\alpha=10$ and the threshold constant $\kappa=0$. The spreading model is fixed to be SI with a transmission parameter $\lambda=1$, on RRGs of average degree $d=3$. Varying $\rho$ on the x-axis, on the main panel we compare the rescaled overlap to the rescaled mean overlap, both at time $t=0$, while on the inset panel we plot the difference between the two. Furthermore, we look at different values of $N$, to study how the behaviour changes when increasing the size of the system. Each point in the plot has been computed by averaging 25 different simulations, and the shaded region represents the $99\%$ confidence interval.
    }\label{fig:Nishi_Rade}
\end{figure}

In Fig.~\ref{fig:Nishi_Rade}, we consider the setting analysed in Sec.~\ref{sec:Rade}, looking at the SI model with $\lambda=1$, and fixing as an example $\alpha=10$ and $\kappa=0$. We compute the values of the rescaled overlap and mean overlap as a function of $\rho$, looking at different sizes $N$. We see that the more we increase $N$, the more the transition gets sharp, signifying that the model presents strong finite size corrections even at moderately large sizes. In the inset plot we look at the difference between  overlap and mean overlap, to investigate the Nishimori conditions. We see that taking $N$ larger shrinks the region where the conditions are not satisfied, leading us to conjecture that in the large $N$ limit, when the transition is very sharp, the Nishimori conditions are satisfied for all values of $\rho$. 
\subsection{\label{App:Ensemble}Other graph ensembles}
In the main text, we have fixed the ensemble of graphs in which the spreading happens to be Random Regular Graphs (RRGs) of degree $d=3$, for convenience. Here we present results for some other variants of locally tree-like random graphs, to show that the phenomenology we describe in the paper remains unchanged. Specifically, we change the degree of the graphs to $d=5$, and we compare the ensemble of RRGs to Erdos-Renyi (ER) graphs~\cite{erdos59a}, where each couple of nodes is randomly connected with probability $p$. To make the two ensembles of graphs comparable, we set $p = \frac{d}{N-1}$, such that in both cases the average degree of the nodes is $d$. Moreover, if the graph is composed by multiple disconnected components, we consider only the biggest one and neglect the others (for $d=5$ and $N$ large enough, the largest component comprehends almost all the nodes with high probability).
\begin{figure}
    \centering
    \includegraphics[width=0.8\linewidth]{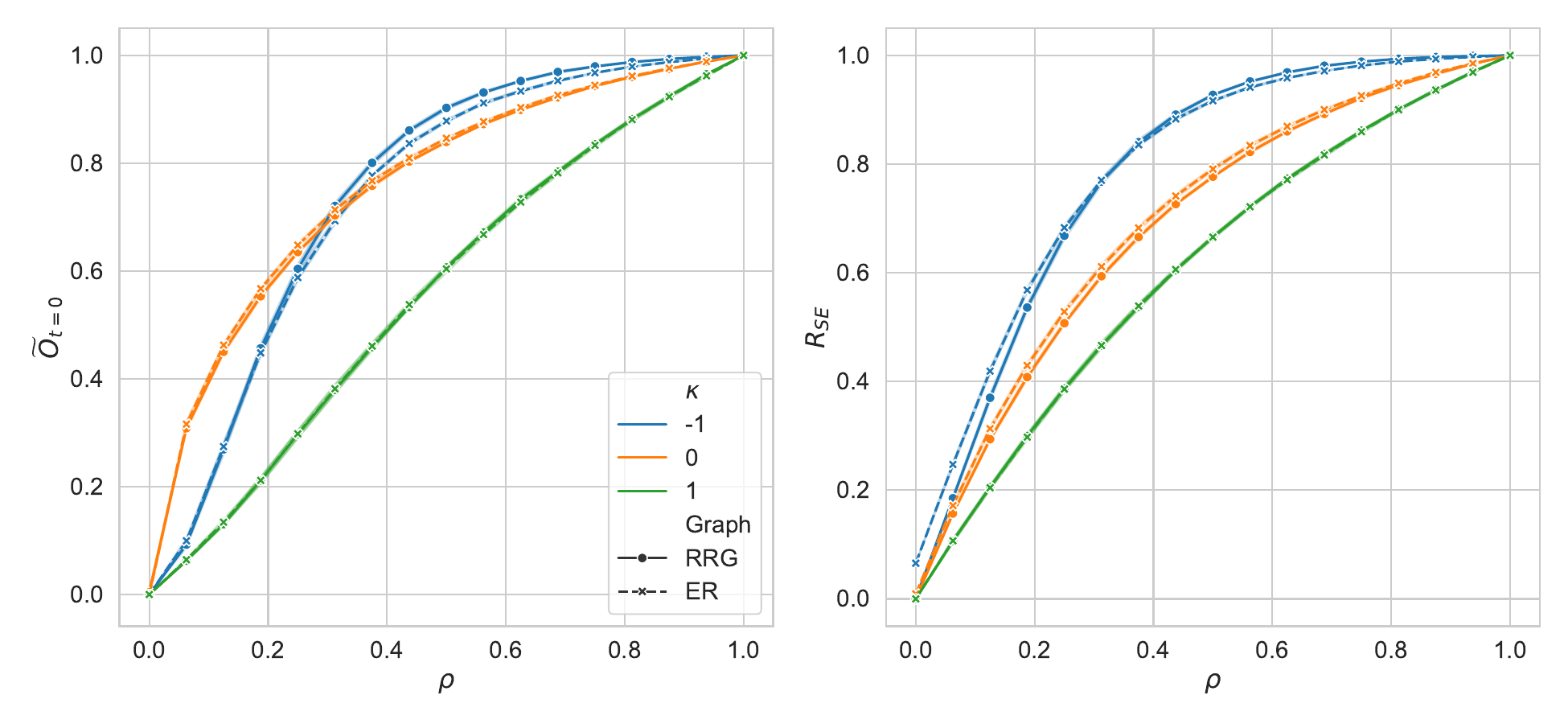}
    \caption{\textbf{Comparison between Ensembles.} We consider the PSS model with Gaussian weights, fixing $\alpha=4$ and changing the threshold constant $\kappa=-1,0,1$, corresponding to an average fraction of sources $\delta \approx 0.841, 0.5, 0.159$ respectively.. The spreading model is fixed to be SI with a transmission parameter $\lambda=1$, and we compare RRGs and ER graphs, fixing the number of nodes $N=20000$ and the average degree $d=5$ as explained in the text. On the left panel we plot the rescaled overlap at time $t=0$ defined in~\eqref{eq:resOv}, while on the right the rescaled squared error defined in~\eqref{eq:res_se}, both as a function of the fraction of sensors $\rho$. Each point in the plot has been computed by averaging 20 different simulations, and the shaded region represents the $99\%$ confidence interval.}\label{fig:ensemble}
\end{figure}

In Fig.~\ref{fig:ensemble} we look at both the overlap at time $t=0$ and the squared error, considering the PSS model with Gaussian weights and varying threshold $\kappa$. We show that the choice of the ensemble impacts only slightly the inference capabilities of our algorithm, which justifies our choice of fixing the graph's ensemble in the main text.

\newpage
\nocite{*}

\bibliographystyle{apsrev4-2}
\bibliography{refs}

\end{document}